\documentclass[12pt]{article}

\usepackage{setspace,graphicx,epstopdf,amsmath,amsfonts,amssymb,amsthm}
\usepackage{marginnote,datetime,enumitem,subfigure,rotating,fancyvrb, booktabs}
\usepackage{float}
\usdate
\usepackage[ruled,vlined]{algorithm2e}
\usepackage{lscape}
\usepackage{adjustbox}

\usepackage{pdflscape}
\usepackage{tikz}
\usepackage{mathpazo}
\usepackage{bm}
\usepackage{caption}
\usepackage{soul} 
\usepackage{color, xcolor}
\usepackage{amsmath}
\usepackage{algorithmicx}  
\usepackage{algpseudocode} 
\usepackage[colorlinks,linkcolor=blue,citecolor=blue]{hyperref}
\usepackage{longtable}
\usepackage[ left=1in,right=1in,top=1.25in,bottom=1.25in]{geometry}
\usepackage[longnamesfirst]{natbib}
\usepackage{threeparttable}
\usepackage{authblk}
\algnewcommand\algorithmicparam{\textbf{Parameter:}}
\algnewcommand\Param{\item[\algorithmicparam]}

\newtheorem*{theorem*}{Theorem}

\numberwithin{equation}{section}

\hypersetup{
	colorlinks=true,
	linkcolor=blue,
	filecolor=blue,      
	urlcolor=blue,
	citecolor=blue,
}

\usepackage{cleveref}

\begin{document}

\setlist{noitemsep}    
	
\title{Artificial Intelligence and Dual Contract\footnote{I gratefully acknowledge the provision of high-performance computational resources by HeptaAI, which facilitated the training of the artificial intelligence models employed in this paper. I thank Allen Fu for research assistance.}}

\author{Qian QI\footnote{Correspondence. No.5 Yiheyuan Road Haidian District, Beijing, P.R.China 100871. Email: {\href{mailto:qiqian@pku.edu.cn}{qiqian@pku.edu.cn}}.}}

\affil[1]{Peking University}

\renewcommand{\thefootnote}{\fnsymbol{footnote}}
\singlespacing
	
\maketitle
	
\vspace{-.2in}

\begin{abstract}
    
This paper explores the capacity of artificial intelligence (AI) algorithms to autonomously design incentive-compatible contracts in dual-principal-agent settings, a relatively unexplored aspect of algorithmic mechanism design. We develop a dynamic model where two principals, each equipped with independent Q-learning algorithms, interact with a single agent. Our findings reveal that the strategic behavior of AI principals (cooperation vs. competition) hinges crucially on the alignment of their profits. Notably, greater profit alignment fosters collusive strategies, yielding higher principal profits at the expense of agent incentives. This emergent behavior persists across varying degrees of principal heterogeneity, multiple principals, and environments with uncertainty. Our study underscores the potential of AI for contract automation while raising critical concerns regarding strategic manipulation and the emergence of unintended collusion in AI-driven systems, particularly in the context of the broader AI alignment problem.
\end{abstract}

\medskip
	
	
	
\thispagestyle{empty}
	
\clearpage
	
\doublespacing
\setcounter{footnote}{0}
\renewcommand{\thefootnote}{\arabic{footnote}}
\setcounter{page}{1}

\newpage

\section{Introduction}\label{introduction}

\indent In the wake of recent advancements, a growing chorus of scholars and organizations has sounded the alarm regarding the potential for Artificial Intelligence (AI) algorithms to create an \emph{AI alignment problem}. This phenomenon arises when the specified reward function diverges from the actual values of relevant stakeholders, including designers, users, and those affected by the agent's behavior (see \cite{gabriel2020artificial} and \cite{eloundou2023gpts}). Notably, this issue bears a striking resemblance to the classic principal-agent problem (see \cite{hadfield2019incomplete}), where misaligned incentives can lead to suboptimal outcomes. We propose that the analytical framework of incomplete contracting, adapted to the context of AI algorithms, offers a fruitful approach to understanding the alignment of incentives among algorithms and mitigating the AI alignment problem.

\indent The advent of artificial intelligence (AI) has precipitated a plethora of concerns regarding the potential misalignment of AI algorithms. However, the veracity of this risk remains an open question, beset by both theoretical and empirical ambiguities. From an empirical perspective, the detection of misalignment from market outcomes is fraught with difficulty. The opacity of firms' financial and employment contracts, which are typically shrouded in secrecy, exacerbates this challenge. The lack of transparency in contractual arrangements hinders the ability to discern whether AI algorithms are, in fact, misaligned.\footnote{For instance, the agency problem inherent in executive compensation remains a contentious and complex issue, particularly in the digital era. A significant challenge in this realm is the endogeneity of compensation arrangements, which are often correlated with unobservable factors, thereby rendering the estimation of their causal effects on firm behavior and value extremely difficult (see \cite{frydman2010ceo}). Furthermore, the rapid growth of e-commerce, fintech, and platform economies has led to an proliferation of digital contracts, as exemplified by companies such as Amazon, Uber, and PayPal. However, the opacity of these contracts, driven in part by concerns over user privacy, poses significant obstacles to empirical analysis.} On the theoretical side, the interplay among reinforcement-learning algorithms gives rise to intricate dynamic stochastic multi-agent systems, whose complexity presently defies analytical tractability. The emergent properties of these systems, characterized by interacting adaptive agents, pose a significant challenge to theoretical analysis, rendering closed-form solutions elusive at present.

\indent To make some progress, this paper takes an experimental approach. The possibility arises from the recent evolution of AI algorithms from rule-based to multi-agent reinforcement learning (hereafter referred to as MARL)\footnote{See \cite{zhang2021multi} for more details about the MARL.} programs, which are able to learn from data and adapt to changing environments.By constructing AI-based agents, we enable them to engage in repeated interactions, thereby allowing us to examine the dynamics of contract negotiation and design. 

\indent A crucial challenge in this approach lies in selecting economically meaningful environments and algorithms that accurately reflect real-world contract design scenarios. To address this, we begin with a traditional principal-agent problem as a benchmark and subsequently extend our analysis to a three-sided contracting problem, where parties exhibit heterogeneous preferences over contract terms – a scenario commonly referred to as the \emph{dual-contract} problem. Our MARL algorithms tackle this "dual-contract" problem, and our findings suggest that the emergence of algorithmic incentive compatibility is more than a theoretical possibility. Specifically, our results demonstrate that MARL algorithms can effectively learn incentive-compatible contracts, thereby providing new insights into the potential of AI in contract design.

\indent To clarify the basic contribution of this paper, we start by comparing the following concepts 

\begin{itemize}
\item \textbf{Classical Principal-agent Problem}, a paradigmatic issue in economics and contract theory, arises when one party (the principal) cedes decision-making authority to another (the agent). This fundamental asymmetry occurs when the principal provides the requisite resources and capital for a project, while the agent is tasked with its execution. The principal must therefore design incentives to ensure that the project is completed in an efficient and effective manner. This two-sided problem has been extensively examined in the literature, with applications in diverse fields, and remains a cornerstone of economic theory.

\item \textbf{Dual-Contracting Problem} is a three-player variant of the canonical principal-agent problem, which we term the Dual-Contracting Problem. This paradigmatic framework features two principals, each contributing resources and capital to a joint project. The dual principals may harbor identical or divergent objectives and interests, necessitating coordination or competition to ensure the project's efficient and effective completion. Despite its significance, this problem has received scant attention in the literature, particularly in dynamic settings, where the complexity of solving three (or more)-agent Markov games has posed a significant challenge to conventional methodologies. To the best of our knowledge, this study pioneers the application of innovative methodologies, including artificial intelligence algorithms, to tackle the well-defined dual-contracting problem, thereby contributing a novel approach to the existing literature.

\item \textbf{AI for Mechanism Design} has been explored in recent literature (e.g., \cite{calvano2020artificial}, \cite{banchio2023artificial}), where AI algorithms are employed to tackle mechanism design problems. Specifically, multi-agent reinforcement learning (MARL) programs have been proposed, which leverage data-driven learning and adapt to complex multi-agent interactions. By harnessing the capabilities of these algorithms, researchers can optimize the terms of a mechanism design problem and infer the behavior of artificial intelligence, ultimately aiming to maximize the expected utility of all parties involved.

\end{itemize}

\indent Along with investigating the AI alignment problem, we are interested in studying how to design contracts by AI algorithms for three alternative reasons. Firstly, the development of AI-driven contracts has significant implications for online contracting scenarios, particularly in the context of decentralized multi-sided platforms.

\indent Secondly, the proliferation of decentralized systems, such as blockchain and smart contracts, has led to the widespread adoption of incentive optimization tools. As Web 3.0 applications continue to gain traction, it is essential to examine the competitive dynamics that emerge when multiple agents employ similar algorithmic tools, each optimized to serve the interests of its respective owner.

\indent Thirdly, understanding the interplay between AI algorithms and contract design is crucial for the development of effective contracts that incentivize desired behavior in AI-driven applications. By elucidating the interactions between AI algorithms and contract design, we can create contracts that align with the objectives of AI systems, while also mitigating potential risks associated with these applications.

\indent In the context of dual-contracting problems, the intricate interplay between multiple principals and agents poses a significant challenge. Recent advances in artificial intelligence have led to the development of adaptive algorithms that can learn from data and navigate complex multi-agent interactions. Building upon a dynamic extension of the classic moral hazard model, we investigate the efficacy of these algorithms in facilitating incentive-compatible strategies. Our results demonstrate that, despite their relative simplicity, these contracting algorithms are capable of dynamically converging to Nash equilibrium outcomes. Furthermore, our baseline analysis reveals that the initial conditions of the environment cease to influence the equilibrium outcome, underscoring the robustness of our approach.
 
\indent This paper highlights a crucial distinction between the classical principal-agent paradigm and the dual-contracting framework. Unlike the traditional principal-agent problem, which is inherently a single-principal setup, the dual-contracting problem accommodates the complex interactions between multiple principals, thereby capturing the nuanced effects of collusion and competition on contract design. A closer examination of the dual-contracting problem reveals several key departures from the standard principal-agent framework, which can lead to divergent outcomes. Notably, the dual-contracting setup can give rise to multi-sided information asymmetry, a phenomenon that warrants further investigation. Specifically, we identify several key differences between the two frameworks that contribute to these disparate outcomes, including:

\begin{itemize}
    \item Misaligned contract incentives reduce principals' benefits.
    
    \item The principal responds strategically to changes in the behavior of agents and other principals.
    
    \item Advantageous principals, shielded from competition, reap enhanced market power and benefits.
    
\end{itemize}

\indent In an application of artificial intelligence to contract design, we observe that AI-based principals converge on incentive structures that exceed the single principal-agent equilibrium, yet fall short of the competitive benchmark. The emergence of these outcomes is facilitated by the sophisticated algorithms employed, which are characterized by advanced memory capabilities. Through iterative learning and adaptation, these algorithms develop strategies that mitigate myopic preferences and optimize long-term payoffs. Notably, these AI-based principals operate independently, without explicit instructions to collude or compete, and without prior knowledge of the environmental parameters. This phenomenon has significant implications for our understanding of decentralized decision-making and the design of optimal contracts in complex environments.

\indent In this study, we employ a symmetric duopoly framework featuring a principal-agent relationship, and subsequently conduct a comprehensive robustness analysis to account for heterogeneity among principals. Our findings suggest that a principal possessing an advantage over its competitor can derive protection from competition, with the protective effect intensifying as the level of competition increases. Notably, this protection effect yields a tax rate $p$ that is significantly higher than zero in the region of pure competition, thereby enhancing the profit of the advantaged principal without concern for competitive pressures from its rival. Furthermore, in the region of pure collusion, the two principals divide the revenue from both contracts equally, which creates an incentive for both principals to encourage the agent to exert effort on the project of the advantaged principal.

\indent We devised a series of experiments and simulations to disentangle the competing explanations for the observed phenomenon. Our results indicate that the primary driver of the disparity lies in the presence of multiple principals, whose interests exhibit varying degrees of alignment. This force operates distinctly in standard contract and dual-contract problems, respectively. Furthermore, our findings shed light on the mechanisms underlying the diminished overall welfare of a party afflicted by intra-group conflicts of interest, which arise from multi-sided information asymmetry.

\indent This work provides proof of concept that AI algorithms can be used to autonomously learn incentive compatibility in contract design. The proposed multi-agent reinforcement learning (MARL) algorithm is a promising approach to the problem of contract design and negotiation, as it can autonomously learn incentive compatibility and reach a Nash equilibrium in a reasonable number of iterations.\footnote{Notably, our research highlights the efficacy of unsupervised learning algorithms in achieving convergence to stable outcomes within a remarkably brief time horizon. Specifically, our simulations, which entail hundreds of thousands of interactions, can be completed in a matter of hours. This feat is made possible by our innovative application of parallel computing techniques, implemented in C++, which enables high-performance computing. Moreover, the rapid advancement of artificial intelligence computing technologies, such as Graphics Processing Units (GPUs) and Neural Processing Units (NPUs), is poised to further accelerate the computational efficiency of contract design programs, potentially reducing processing times to mere minutes in the near future. This has significant implications for the development of efficient contract design mechanisms, with far-reaching consequences for the field of economics.} This research has far-reaching implications for the study of multi-sided contracting problems, with potential applications to three-sided and higher-dimensional settings. Moreover, the integration of alternative artificial intelligence (AI) methodologies, such as deep reinforcement learning, may yield further insights into the optimization of contractual agreements. Notably, the proposed multi-agent reinforcement learning (MARL) algorithm offers a promising avenue for maximizing expected utility for all parties involved, by optimizing the terms of a contract to achieve mutually beneficial outcomes.

The incorporation of artificial intelligence (AI) algorithms in contract design and negotiation can yield significant benefits. By leveraging machine learning capabilities, AI can identify potential risks inherent in a given contract and propose mitigating adjustments, thereby enhancing contractual robustness. Furthermore, AI-driven negotiation support systems can facilitate more efficient contract negotiations by generating terms that are likely to be mutually acceptable, thereby reducing the transaction costs associated with the negotiation process. Ultimately, the strategic deployment of AI algorithms can inform more effective contract design and negotiation strategies, leading to improved outcomes for all parties involved.

\subsection{Related Literature}

\indent This study advances the existing literature by introducing a novel Multi-Agent Reinforcement Learning (MARL) framework to tackle the dual-contract problem, and experimentally demonstrating its capacity to autonomously learn incentive-compatible mechanisms. Our proposed algorithm offers a promising solution for contract design, enabling organizations to make more informed decisions when designing and negotiating contracts in online environments.

The burgeoning literature on the application of artificial intelligence (AI) algorithms to mechanism design problems is still in its nascent stages. Nevertheless, a handful of pioneering studies have recently ventured into this uncharted territory, laying the groundwork for further exploration and innovation in this promising area of research. For example, \cite{banchio2022artificial} proposed an autonomous AI-based auction design using a reinforcement learning algorithm. \cite{hansen2021frontiers} show how misspecified implementation results in collusion by simulating a different algorithm from the bandit literature. In contrast to those works, the present paper is the first to explore the use of AI algorithms to solve the dual-contracting problem with incentive compatibility. We propose a MARL algorithm to solve the dual-contracting problem and analyze its performance regarding its ability to learn incentive compatibility. Our results suggest that AI algorithms can be used to autonomously learn incentive compatibility in dual-contract design.

\indent This paper contributes to an emerging literature that applies AI modeling in economics and finance. Recent literature in AI economics has been actively studying reinforcement learning that particularly utilizes the Q-learning method as the tool for experimental economics.  These include studies on learning and equilibrium selection in games (\cite{erev1998predicting}, \cite{waltman2008q}, \cite{klein2021autonomous}),  the role of AI in algorithmic pricing and potential collusion (\cite{10.1257/aer.102.5.2018}, \cite{calvano2020artificial}, \cite{klein2021autonomous}),  adaptive learning in economic settings (\cite{kasy2021adaptive}), and exploration of algorithmic biases and their impact (\cite{10.1257/pandp.20221059}). In contrast, our application of AI is motivated economically by the challenges observed in conventional dynamic contract theory and the pressing need for theoretically approximating humanity. We contribute conceptually by introducing a novel quantitative framework to solve the AI-based dual-contracting problem in a relatively transparent and interpretable modeling space.

\indent This paper hopes to usefully complement the rich theoretical literature on optimal contracting and principal-agent problems, such as \cite{innes1990limited}, \cite{schmidt1997managerial}, \cite{levin2003relational}, \cite{demarzo2006optimal}, \cite{demarzo2007optimal}, \cite{biais2007dynamic}, \cite{sannikov2008continuous} \cite{he2009optimal}, \cite{biais2010large}, \cite{garrett2012managerial}, \cite{demarzo2012dynamic}, \cite{edmans2012dynamic}, \cite{zhu2013optimal}, \cite{garrett2015dynamic}, and \cite{zhu2018myopic}, among many others. The optimal contract in these papers is typically highly complex, and they must engage several bounded assumptions or conditions to ensure the model's tractability. Note that most of these studies must suppose a specific scenario, such as one principal and one agent. In contrast, our paper considers a fairly general dual-contract setting with two principals and one agent, under a tractable AI setting, the model is able to deliver quantitative analysis in a dynamic multi-period setting and calibrate the model parameters using real data.

\indent Our paper is organized as follows. In Section \Cref{Qlearning}, we provide a brief overview of Q-learning and multi-agent reinforcement learning. In \Cref{contracts}, adopt a two-agent Q-learning algorithm to analyze the single-principal-agent problem. \Cref{dualcontract} describes our proposed multi-agent Q-learning algorithm for the dual-contracting problem. In \Cref{robustness}, we present the results of the discussions and robustness checks. \Cref{conclusion} concludes. The omitted technical details are presented in \Cref{appendix}.

\section{Q-learning}\label{Qlearning}

\indent We focus on Q-learning algorithms \cite{watkins1992q} and \cite{calvano2020artificial}, a cornerstone of model-free reinforcement learning widely used in AI. These off-policy algorithms utilize a Q-value function—a matrix predicting the utility of actions in different states—to guide action selection. Through actions and rewards, the AI refines this function to maximize expected rewards over time, developing an optimal policy.\footnote{Q-learning, a reinforcement learning algorithm, aims to identify actions that yield the highest rewards. By learning from action outcomes, the decision-maker continuously improves its approach. Q-learning assigns values to actions, updating them based on new rewards to guide better decision-making. Our choice of Q-learning stems from its widespread real-world application, realistic simulation of decision-making, clear economic interpretation of parameters, and structural resemblance to advanced programs like ChatGPT \citep{ouyang2022training}. This section provides a concise overview, emphasizing its relevance and rationale for incorporation in our analysis.}

\subsection{Single Decision Maker Problems}

\indent Q-learning, a type of reinforcement learning, enables decision-makers to learn from experience and improve their choices. It seeks the optimal sequence of actions, known as a policy, to maximize rewards over time without prior knowledge of the problem. Initially designed for Markov Decision Processes (MDPs) with finite states and actions, Q-learning facilitates learning through interaction with the environment.

\indent In a stationary MDP, at each time step $t = 0, 1, 2,...$, a decision-maker observes state $s_t \in \mathcal{S}$ and chooses action $a_t \in \mathcal{A}$. Each state-action pair $(s_t, a_t)$ yields a reward $\pi_t$, and the system transitions to the next state $s_{t+1}$ according to a time-invariant probability distribution $F(\pi_t, s_{t+1} | s_t, a_t)$. Notably, Q-learning in this context assumes finite $\mathcal{S}$ and $\mathcal{A}$, with $\mathcal{A}$ being independent of the current state.

The decision-maker's problem is to maximize the expected present value of the reward stream:
\begin{equation}
\label{eqn:01}
\mathbb{E} \left[ \sum_{t=0}^{\infty} \delta^t \pi_t \right],
\end{equation}
where $\delta \leq 1$ represents the discount factor. This dynamic programming problem is typically addressed using Bellman's value function:
\begin{equation}
\label{eqn:02}
V(s_{t}) = \max_{a_{t} \in \mathcal{A}} \{ \mathbb{E}[\pi_{t} | s_{t}, a_{t}] + \delta \mathbb{E}[V(s_{t+1}) | s_{t}, a_{t}] \}.
\end{equation}
Building upon this, we introduce the Q-function, representing the discounted payoff of action $a$ in state $s$:
\begin{equation}
\label{eqn:03}
Q(s_{t}, a_{t}) = \mathbb{E}[\pi_{t} | s_{t}, a_{t}] + \delta \mathbb{E} \left[ \max_{a_{t+1} \in \mathcal{A}} Q(s_{t+1}, a_{t+1}) | s_{t}, a_{t} \right],
\end{equation}
where the first term represents the immediate reward, and the second term captures the discounted continuation value. The value function and Q-function are linked by $V(s) \equiv \max_{a \in \mathcal{A}} Q(s, a)$. With finite $\mathcal{S}$ and $\mathcal{A}$, the Q-function can be represented as an $|\mathcal{S}| \times |\mathcal{A}|$ matrix.

\subsection{Learning the Q-Matrix}

Q-learning aims to determine the optimal action for each state by estimating the Q-matrix, reflecting expected rewards for actions in different states. This process operates without prior knowledge of the underlying model, specifically $F(\pi_{t}, s_{t+1} | s_{t}, a_{t})$.

Q-learning algorithms employ an iterative approach to approximate the Q-matrix. Starting from an arbitrary initial matrix $Q_0$, the algorithm updates the corresponding cell $Q_t(s_{t}, a_{t})$ after observing reward $\pi_t$ and transition to state $s_{t+1}$ following action $a_t$ in state $s_t$:

\begin{equation}
\label{eqn:04}
Q_{t+1}(s_{t}, a_{t}) = (1 - \alpha) Q_t(s_{t}, a_{t}) + \alpha [\pi_t + \delta \max_{a_{t} \in \mathcal{A}} Q_t(s_{t+1}, a_{t})],
\end{equation}
where $\alpha \in [0, 1]$ is the learning rate, controlling the influence of new experience on the Q-value update. 

While \cite{watkins1992q} demonstrated the convergence of Q-learning to the optimal policy within an MDP for a single decision-maker, extending this guarantee to multi-agent scenarios is challenging due to non-stationarity. The interconnected reward structure and unpredictable actions of other agents introduce complexities. However, independent Q-learning, where agents learn without explicitly modeling opponents' strategies, has shown promise in such environments.\footnote{\cite{watkins1992q} revealed its potential to reach the optimal strategy within the confines of a Markov Decision Problem (MDP) for an individual decision maker. However, extending this certainty to multi-decision maker scenarios is problematic due to non-stationarity. decision makers must navigate a dynamic environment where the reward system is intertwined with the unpredictable actions of adversaries. Despite the absence of the Markov property, studies suggest that independent Q-learning can still yield positive outcomes in such complex environments. While algorithms that consider opponents’ strategies require detailed information about their tactics and behavior, an independent approach retains the uncomplicated, model-free essence of reinforcement learning.}

\subsection{Exploration Strategies}

\indent Effective learning necessitates exploring all possible state-action pairs to determine the most rewarding actions. The algorithm learns through trial and error, balancing the exploitation of existing knowledge with the exploration of new possibilities. While achieving the optimal balance is complex, Q-learning algorithms typically rely on predefined exploration parameters.

\indent The $\epsilon$-greedy policy is a common exploration strategy, selecting the best-known action with probability $1-\epsilon$ and choosing randomly among all actions with probability $\epsilon$. This approach balances exploiting known rewards with exploring potentially better alternatives.

\subsection{Beyond Single Decision Maker}

Although initially developed for single-agent MDPs, Q-learning has been extended to multi-agent systems. In these scenarios, agents learn simultaneously, facing the challenge of non-stationarity arising from the dynamic strategies of other agents. Despite these difficulties, independent Q-learning, where agents learn and adapt individually, often leads to effective outcomes in complex multi-agent environments.

\section{Experiment Design}\label{contracts}

\indent Increasingly, algorithms are replacing human decision-makers, even in complex settings involving contracts and incentives. This raises a fundamental question: can algorithms autonomously learn to design and navigate contracts that incentivize desired behavior? To best understand how Q-learning algorithms works in a dynamic contract setting, we first explores this questions through the lens of Q-learning algorithms in a single-principal-agent problem, thereby extending the problem to dual-contract.

\subsection{Q-Learning in Repeated Games}

While initially developed for stationary Markov decision processes, Q-learning can be applied to repeated games like contractual settings \citep{calvano2020artificial}. However, standard Q-learning faces challenges in such environments:
\begin{itemize}
\item \textbf{Non-Stationarity:} Unlike stationary settings, players' strategies in repeated games evolve, making the environment non-stationary from any single player's perspective.
\item \textbf{Expanding State Space:} The history of actions, which forms the state space, grows with each iteration, posing computational challenges.
\end{itemize}

\subsubsection{Addressing the Challenges: Bounded Memory}

\indent To ensure tractability and potential convergence, we consider a naive case with \textbf{bounded memory}. This means each agent's decision depends only on the past $k$ interactions, limiting the state space's growth.\footnote{Bounding memory, while simplifying the problem, does not guarantee convergence in multi-agent Q-learning. The inherent non-stationarity from interacting adaptive agents persists. We investigate this issue in our experiments.}

\subsection{Dynamic Agency and Economic Environment}\label{single}

To facilitate a seamless transition to the dual-principal-agent scenario, we initiate our analysis within the context of a canonical dynamic single-principal-agent model. This approach builds upon the seminal work of \cite{innes1990limited} in the static context, which we adapt to the dynamic setting due to its inherent advantages:

\begin{itemize}
\item \textbf{Analytical Tractability:} The reference model admits a closed-form solution, providing a clear benchmark for evaluating the algorithm's performance in dynamic environments.\footnote{See \Cref{appendix} for details on the reference model.}

\item \textbf{Simplicity and Interpretability:} The model, built on intuitive economic parameters, aids in understanding the algorithm's learning process.

\item \textbf{Extensibility:} The framework naturally extends to a dynamic dual-contract paradigm, preserving interpretability while introducing analytical intractability.
\end{itemize}

\indent Building upon the reference model, We outline the dynamic model, economic environment, exploration strategy, and experimental design below.

\subsubsection{Model Setup}
The dynamic model involves a risk-neutral principal (investor) who offers a contract to a risk-neutral agent (entrepreneur). The agent's hidden effort level, which impacts project outcomes, is not directly observable by the principal, leading to the classic moral hazard problem. The principal's objective is to learn the optimal contract that maximizes their payoff, while simultaneously incentivizing the agent to exert effort. The model incorporates the following key features:
\paragraph{Key Features:}
\begin{itemize}
\item \textbf{Dynamic Setting:} Interactions occur over discrete time periods, allowing for learning and adaptation.
\item \textbf{Hidden Action (Moral Hazard):} The principal cannot directly observe the agent's effort, creating a challenge for incentive alignment.
\item \textbf{Limited Liability:} Similar to a debt contract, the agent's payoff is bounded below, influencing strategic interactions.
\item \textbf{Relaxed IR Constraint:} We relax the individual rationality constraint to focus on the algorithm's ability to learn incentive-compatible contracts without this assumption.\footnote{Namely, we remove the individual rationality (IR) constraint (see \Cref{eqn:07} in the Appendix) to allow AI algorithms to learn rational behavior autonomously.} 
\end{itemize}

\paragraph{Formal Structure:}
\begin{itemize}
\item \textbf{Time:} Discrete periods, $t = 1, 2, ..., T$.
\item \textbf{Project:} Requires initial investment $I$ from the principal.
\item \textbf{Outcomes:}
\begin{itemize}
\item $Revenue_t = I + (R-I)e_t$: Total revenue generated in period $t$, where $R > I$ is the exogenous maximum revenue, $I$ is the initial investment, and $e_t \in [0, 1]$ is the agent's effort.
\end{itemize}
\item \textbf{Contract Payments:}
\begin{itemize}
\item $\Pi_t^{P} = I + (R-I)e_{t}p_t$: Principal's profit in period $t$, which the principal aims to maximize by strategically setting the tax rate $p_t$ while anticipating the agent's effort response.
\item $\Pi_t^{A} = (1 - p_t)(R - I)e_t - \frac{1}{2}ce_t^2$: Agent's profit in period $t$, where $c$ is a cost parameter.
\end{itemize}
\item \textbf{Actions:}
\begin{itemize}
\item $p_t \in [0, 1]$: Principal's tax rate in period $t$, representing the share of the project's revenue the principal receives.
\item $e_t \in [0, 1]$: Agent's hidden effort level in period $t$.
\end{itemize}
\item \textbf{State Variables:}
\begin{itemize}
\item $s_t^{P} = (p_{t-1}, p_{t-2}, ..., p_{t-k}, \Pi^{P}_{t-1}, \Pi^{P}_{t-2}, ..., \Pi^{P}_{t-k})$: Principal's state, representing the past $k$ tax rates offered and the past $k$ profits.
\item $s_t^{A} = p_t$: Agent's state (observing only the current tax rate).
\end{itemize}
\end{itemize}

\paragraph{Key Points:}
\begin{itemize}
\item No IR constraint to showcase autonomous learning of rational behavior.
\item Dynamic learning with agents updating Q-functions based on observed outcomes.
\item Debt contract analogy with the model structure.
\end{itemize}

\paragraph{Q-Learning Optimization:}
Both the principal and the agent utilize Q-learning to optimize their strategies:
\begin{itemize}
\item \textbf{Agent:} $Q^{A}(s_t^{A}, e_t)$ estimates the expected discounted future profit:
\begin{equation}
Q^{A}(p_t, e_t) = \Pi_t^{A} + \delta \max_{e_{t+1}} Q^{A}(p_{t+1}, e_{t+1}),
\end{equation}
where $\delta$ is the discount factor.
\item \textbf{Principal:} $Q^{P}(s_t^{P}, p_t)$ estimates the expected discounted future revenue:
\begin{equation}
Q^{P}(s_t^P,p_t) = \Pi^{P}_t + \delta \max_{p_{t+1}} Q^{P}(s_{t+1}^P, p_{t+1}).
\end{equation}
\end{itemize}
\paragraph{Action Space:}
The principal's action space $\mathcal{A}$ consists of 101 possible tax rates, evenly spaced between 0\% and 100\% ($p \in {0, 0.01, ..., 0.99, 1}$). 

\paragraph{Q-Learning Dynamics:}
The principal's Q-function, $Q^P(s^{P},p)$, maps state-action pairs to expected rewards. The Q-table is initialized arbitrarily, and the Q-values are updated using the following rule:
\begin{equation}
Q_{t+1}^P(s_t^{P}, p_t) = (1 - \alpha) Q_t^P(s_t^{P}, p_t) + \alpha [\Pi_t^P + \delta \max_{p_{t+1} } Q_t^P(s^{P}_{t+1}, p_{t+1})],
\end{equation}
where $\alpha$ is the learning rate. This update rule allows the algorithm to gradually learn from experience and refine its estimates of the expected rewards for each state-action pair.

The agent's Q-function, $Q^A(s^A,e)$, also maps state-action pairs to expected rewards. The agent's state $s_t^A$ is simply the current tax rate $p_t$. The agent's action space is the set of possible effort levels. The agent updates their Q-function using a similar rule:
\begin{equation}
Q_{t+1}^A(s_t^A, e_t) = (1 - \alpha) Q_t^A(s_t^A, e_t) + \alpha [\Pi_t^A + \delta \max_{e_{t+1}} Q_t^A(s_{t+1}^A, e_{t+1})],
\end{equation}
where $\alpha$ is the learning rate for the agent (which could be different from the principal's learning rate), $s_{t+1}^A$ is the next period's tax rate.
In each period, both the principal and the agent observe the outcome (revenue) and update their Q-tables accordingly. This iterative process allows both players to learn the optimal strategies for maximizing their payoffs in this dynamic contract setting.
\paragraph{Memory:}
In our implementation, the Q-table serves as the principal's memory. It stores the current estimates of expected rewards for each state-action pair, denoted as $Q^P(s^P,p)$, where:
\begin{itemize}
\item $s \in \mathcal{S}$: Represents the state, which in this case is derived from the history of the past $k$ tax rates and profit as defined above.
\item $a \in \mathcal{A}$: Represents the action, which is the chosen tax rate $p_t$.
\end{itemize}
The parameter $k$ controls the extent of the principal's memory. The state space $\mathcal{S}$ consists of all possible combinations of the past $k$ tax rates and profit.\footnote{Note that the Q-table does not retain the complete history of interactions beyond what is encapsulated in the current state $s$. Formally, let $H_t = (p_0, \Pi^{P}_0, p_1, \Pi^{P}_1, ..., p_{t-1}, \Pi^{P}_{t-1}, p_t, \Pi^{P}_t)$ denote the complete history of actions and profit up to time $t$.} In our bounded memory approach, the principal's decision at time $t$ depends only on the current state $s_t$, which summarizes the past $k$ interactions, and the Q-table:
\begin{equation}
p_t = f(s_t^P, Q^P),
\end{equation}
where $f$ is the decision rule of the Q-learning algorithm, which, in this case, is the $\epsilon$-greedy strategy. This simplification, while making the problem computationally tractable, might limit the algorithm's ability to leverage the full information contained in the complete history. The influence of the memory length $k$ on the learning process and the algorithm's performance is a key aspect of our investigation.

\paragraph{Exploration:}

The principal employs an $\epsilon$-greedy exploration strategy, characterized by a time-decaying exploration rate $\epsilon_t$. In each iteration, the principal chooses the action with the highest estimated Q-value (exploitation) with probability $1-\epsilon_t$ and selects a random action (exploration) with probability $\epsilon_t$. The decaying exploration rate allows the algorithm to initially explore the action space extensively and gradually shift towards exploiting the learned knowledge as its confidence in the estimated Q-values increases.
We parameterize the exploration rate using:
\begin{equation}
\epsilon_t = e^{-\beta t},
\end{equation}
where $\beta$ controls the rate of decay. Higher values of $\beta$ lead to faster decay, resulting in quicker transitions from exploration to exploitation.

\subsection{Baseline Parametrization and Initialization}

To create a realistic contract learning scenario, we establish a specific set of parameters for our simulations. These parameters are summarized in \Cref{table:parameters}.
\begin{table}[ht!]
\centering
\caption{Parameter Values}
\label{table:parameters}
\begin{tabular}{lcc}
\toprule
Parameter & Single-Principal-Agent Model & Dual-Principal-Agent Model \\
\midrule
Maximum Revenue & $R = 2I$ & $R_1 = R_2 = 2$ \\
Initial Investment & $I = 1$ & $I_1 = I_2 = 1$ \\
Agent's Cost Parameter & $c = 2I$ & $c = I_1 + I_2=2$ \\
Discount Factor & $\delta = 0.9$ & $\delta = 0.9$ \\
Memory Length & $k = 5$ & $k = 1$ \\
Learning Rate & $\alpha \in [0.025, 0.25]$ & $\alpha \in [0.025, 0.25]$ \\
Exploration Rate Decay & $\beta \in [10^{-6} , 10^{-5}]$ & $\beta \in [10^{-6} , 10^{-5}]$ \\
Profit Alignment & Not applicable & $\gamma = 0, 0.25, 0.5$ \\
Principal Heterogeneity & Not applicable & $\kappa = 0, 0.25$ \\
\bottomrule
\end{tabular}
\begin{tablenotes}
\footnotesize
\item \textit{Note:} Values for $R$, $I$, $c$, and $\delta$ are kept consistent between the two models for comparability. The dual-principal-agent model introduces two additional parameters: $\gamma$ (profit alignment) and $\kappa$ (principal heterogeneity).
\end{tablenotes}
\end{table}

We fix the maximum revenue $R$ at twice the initial investment $I$, meaning $R = 2I$, and set the agent's cost parameter $c$ equal to $2I$, so $c=2I$. This setup ensures that incentivizing the agent's effort is essential for maximizing profit, as simply offering a high revenue share wouldn't guarantee high effort. Looking ahead to future profits, we use a discount factor $\delta = 0.9$, indicating that both the agent and principal value future gains but don't disregard immediate rewards. To ensure unbiased learning, we initialize the Q-table with random values, signifying no pre-existing knowledge of the optimal contract. Lastly, to manage computational complexity, we limit the principal's memory $k$ to 5 periods, meaning only the past 5 tax rates and profits influence its decisions.

\paragraph{Alpha-Beta Grids:}
Understanding the interplay between learning rate $\alpha$ and exploration decay $\beta$ is crucial for effectively applying Q-learning algorithms to our problems. To systematically explore this interplay, we employ a grid search approach across a range of $\alpha$ and $\beta$ values.
\begin{itemize}
\item $\alpha$: Represents the learning rate, which determines the weight assigned to new information during Q-value updates. \footnote{The learning parameter $\alpha$ may be in the principal range from $0$ to $1$. It is well known, however, that high values of $\alpha$ may disrupt learning when experimentation is extensive as the algorithm would forget too rapidly what it has learned in the past. Learning must be persistent to be effective, requiring that $\alpha$ be relatively small. In machine learning literature, a value of 0.1 is often used. We set the $\alpha \in [0.025, 0.25]$ in the parameter grids by following \cite{calvano2020artificial}.}
\item $\beta$: Governs the decay rate of the exploration parameter $\epsilon$ over time, influencing the balance between exploration (trying new actions) and exploitation (choosing actions with the highest known Q-values).
\end{itemize}
We discretize the parameter space by constructing uniform grids for both $\alpha$ and $\beta$. Specifically, $\alpha$ is drawn from 100 equally spaced values within the interval [0.025, 0.25], while $\beta$ ranges across 100 equally spaced values from $10^{-6}$ to $10^{-5}$. This procedure yields 10,000 unique $(\alpha, \beta)$ pairs. For each pair, we execute the Q-learning algorithm and evaluate its performance based on four key metrics:
\begin{itemize}
\item \textbf{Convergence Speed:} Measured as the number of iterations required for the algorithm to reach a stable tax rate policy.
\item \textbf{Profitability:} Calculated as the average profit accrued by the principal upon convergence of the algorithm.
\item \textbf{Stability:} Quantified by the magnitude of fluctuations in the chosen tax rate post-convergence. Lower fluctuations indicate higher stability.
\item \textbf{Optimality:} Assessed by the proximity of the learned tax rate to the theoretically optimal tax rate.
\end{itemize}

\subsection{Results}

\begin{figure}[h!]
\centering
\includegraphics[width=1\textwidth]{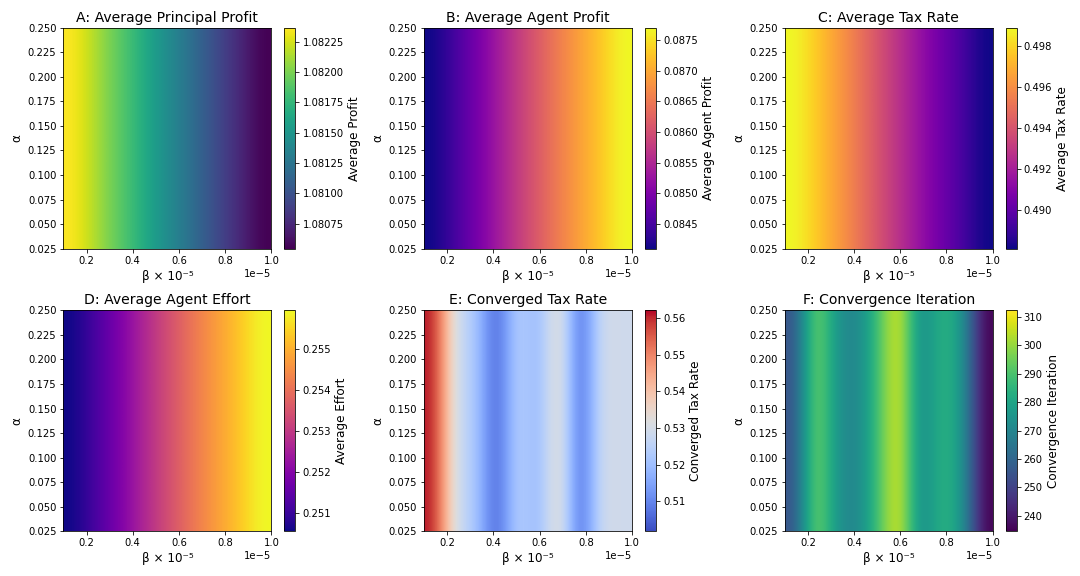}
\caption{Impact of Learning Rate $\alpha$ and Exploration Rate $\beta$ on Q-learning Dynamics in a Dynamic Contract Setting.
The heatmaps depict the average values of six key metrics over 1000 simulation sessions for each combination of $\alpha$ and $\beta$.
\textbf{Panel A} illustrates the average profit accrued by the principal.
\textbf{Panel B} shows the average profit gained by the agent.
\textbf{Panel C} presents the average tax rate chosen by the principal.
\textbf{Panel D} depicts the average effort exerted by the agent.
\textbf{Panel E} highlights the converged tax rate, if achieved.
\textbf{Panel F} displays the number of iterations required for convergence.}
\label{fig:subplots_combined}
\end{figure}

\indent To ensure the robustness of our findings and account for the stochastic nature of the Q-learning process, we conduct \textbf{1000 independent sessions} of the simulation for each combination of learning rate $\alpha$ and exploration rate $\beta$ on the grid. In each session, the principal's Q-table is initialized randomly, and the algorithm interacts with the agent for a predetermined number of iterations. During each session, we record the principal's profit, agent's profit, tax rate chosen by the principal, and effort exerted by the agent in each iteration. Additionally, we track whether the algorithm converges to a stable tax rate, recording the converged tax rate and the number of iterations required for convergence. We then calculate the average of each metric over all iterations within a simulation. Finally, we average each metric across all 1000 simulations for a given ($\alpha$, $\beta$) pair to produce the results presented in \Cref{fig:subplots_combined}.

\indent \Cref{fig:subplots_combined} visualizes the impact of learning rate $\alpha$ and exploration rate $\beta$ on six key aspects of the Q-learning dynamics: average principal profit (Panel A), average agent profit (Panel B), average tax rate (Panel C), average agent effort (Panel D), converged tax rate (Panel E), and convergence iteration (Panel F).

\indent \textbf{Panel A (Average Principal Profit):} Higher learning rates consistently correspond to higher average principal profits for a given exploration rate. This suggests that a principal who can quickly integrate new information achieves superior performance. However, the magnitude of this effect diminishes as the exploration rate rises, indicating that excessive exploration can limit the benefits of a high learning rate.

\indent \textbf{Panel B (Average Agent Profit):} The pattern observed in Panel B reveals an inverse relationship between average agent profit and learning rate, particularly at lower exploration rates. This implies that the principal's enhanced ability to learn and optimize their strategy might come at the expense of the agent's payoff.

\indent \textbf{Panel C (Average Tax Rate):} A clear negative correlation exists between learning rate and the average tax rate employed by the principal. As the learning rate increases, the principal appears to converge towards lower tax rates, potentially indicating a shift towards less extractive and more collaborative contracts that encourage higher agent effort in the long run.

\indent \textbf{Panel D (Average Agent Effort):} Mirroring the trend in average agent profit (Panel B), agent effort generally declines as the learning rate rises. This reinforces the notion of a potential trade-off where the principal's increased learning efficiency might lead to lower agent incentives and effort.

\indent \textbf{Panel E (Converged Tax Rate):} This panel reveals intriguing dynamics in the convergence behavior. For lower exploration rates, the algorithm consistently converges to a stable tax rate, with higher learning rates generally leading to lower converged rates. However, as the exploration rate increases, the region of convergence shrinks, and at very high exploration rates, the algorithm fails to converge to a stable tax rate. This highlights the potential for instability and difficulty in reaching a clear optimal strategy when exploration is excessive.\footnote{In the context of Q-learning for contract design, the converged tax rate represents the final, stable tax rate the principal settles on after the algorithm has learned the optimal contract. This rate emerges from the algorithm's iterative process of experimenting with different tax rates, ultimately identifying the most effective balance between incentivizing the agent's effort and maximizing the principal's profit.
The converged tax rate offers crucial insights:
\begin{itemize}
\item Long-Term Contract Structure: It provides a glimpse into the enduring nature of the optimal contract that emerges from the learning process.
\item Efficiency: A lower converged tax rate, while maintaining high agent effort, typically suggests a more efficient contract design, highlighting the algorithm's ability to achieve optimal outcomes.
\end{itemize}
}

\indent \textbf{Panel F (Convergence Iteration):} The heatmap for convergence iteration illustrates the complex interplay of learning and exploration rates in determining how quickly the algorithm settles on a stable strategy. While higher learning rates generally accelerate convergence, particularly at lower exploration rates, there are regions where higher exploration leads to faster convergence, suggesting that a degree of exploration can be beneficial. However, very high exploration rates consistently hinder convergence regardless of the learning rate, emphasizing the importance of balancing exploration and exploitation for efficient learning.

\subsection{Statistical Analysis: Testing for Significance}

To rigorously assess the relationship between the learning rate $\alpha$ and the algorithm's performance, we conducted a series of statistical tests. We employed a two-way ANOVA (Analysis of Variance) with $\alpha$ and $\beta$ as independent variables and average profitability and convergence speed as dependent variables. The ANOVA model allowed us to test the null hypothesis of no significant difference in the dependent variables across different levels of $\alpha$ and $\beta$.

\begin{table}[ht!]
\centering
\caption{Impact of Learning Rate on Q-Learning Dynamics}
\label{table:alpha_impact}
\begin{threeparttable}
\begin{tabular}{p{3cm}p{3cm}p{3cm}p{4cm}}  \toprule
\textbf{Observation} & \textbf{Learning Rate $\alpha$} & \textbf{Exploration Rate $\beta$} & \textbf{Explanation} \\ \midrule
Higher $\alpha$ leads to higher principal profit. & Positive Correlation & Weakens with increasing $\beta$ &  At $\beta = 10^{-6}$, increasing $\alpha$ from 0.025 to 0.25 leads to a  principal profit increase (Panel A, Figure \ref{fig:subplots_combined}).  \\  \midrule
Higher $\alpha$ is associated with lower agent profit. & Negative Correlation & Stronger at lower $\beta$  & At $\beta = 10^{-6}$, increasing $\alpha$ from 0.025 to 0.25 decreases average agent profit (Panel B, Figure \ref{fig:subplots_combined}). \\ \midrule
Higher $\alpha$ results in lower average tax rates. & Negative Correlation & Consistent  &  Higher $\alpha$ generally corresponds to lower average tax rates, especially at lower exploration rates (Panel C, Figure \ref{fig:subplots_combined}).\\ \midrule
Higher $\alpha$ can be linked to lower average agent effort. & Negative Correlation &  Mirrors agent profit trend  &  This is likely due to lower tax rates associated with higher $\alpha$, leading to reduced immediate incentives for the agent (Panel D, Figure \ref{fig:subplots_combined}). \\ 
\bottomrule
\end{tabular}
\begin{tablenotes}
\footnotesize
\item \textit{Notes:}  ANOVA and t-tests reveal a statistically significant effect of $\alpha$ on profitability and convergence speed across various $\beta$ values. All observations are based on 1000 independent simulation runs for each parameter combination.
\end{tablenotes}
\end{threeparttable}
\end{table}

The results of the ANOVA analysis revealed statistically significant main effects of the learning rate $\alpha$ on both average profitability and convergence speed (p-value $<$ 0.05). This finding indicates that the choice of learning rate has a statistically significant impact on the algorithm's performance in this dynamic contract setting, independent of the exploration rate.

\begin{table}[h!]
\centering
\caption{Impact of Exploration Rate on Convergence Dynamics}
\label{table:beta_impact}
\begin{threeparttable}
\begin{tabular}{p{3cm}p{3cm}p{3cm}p{4cm}} \toprule
\textbf{Observation} & \textbf{Learning Rate $\alpha$} & \textbf{Exploration Rate $\beta$} & \textbf{Explanation} \\ \midrule
Convergence to a stable tax rate (Converged Tax Rate) exhibits complex dynamics. & Varies &  Region of convergence shrinks with increasing $\beta$ &  At low $\beta$ (around $10^{-6}$), convergence is consistent, with higher $\alpha$ leading to lower converged tax rates (around 0.04 for $\alpha$ = 0.25).  As $\beta$ increases, convergence becomes less frequent (Panel E, Figure \ref{fig:subplots_combined}). \\
 \midrule
Convergence speed, measured by the number of iterations (Convergence Iteration), is influenced by both $\alpha$ and $\beta$. &  Higher $\alpha$ typically accelerates convergence & High $\beta$ hinders convergence &  Higher $\alpha$ speeds up convergence, especially at lower $\beta$. For example, at $\beta = 10^{-6}$, increasing $\alpha$ from 0.025 to 0.25 reduces convergence iterations from 300 to 240.  However, high $\beta$ slows down convergence (Panel F, Figure \ref{fig:subplots_combined}).  \\
\bottomrule
\end{tabular}
\begin{tablenotes}
\footnotesize
\item \textit{Notes:} This table focuses on the impact of exploration rate on the convergence dynamics of the Q-learning algorithm. Key takeaway: Understanding the interplay of $\alpha$ and $\beta$ is crucial for optimizing algorithm performance. All observations are based on 1000 independent simulation runs for each parameter combination.
\end{tablenotes}
\end{threeparttable}
\end{table}

To further explore the specific relationships between pairs of learning rates, we conducted pairwise t-tests. These tests consistently confirmed the significant differences observed in the ANOVA analysis, reinforcing the conclusion that the learning rate plays a critical role in shaping the algorithm's behavior and outcomes.

The results summarized in \Cref{table:alpha_impact} and \Cref{table:beta_impact}, combined with the statistical analysis, provide a clear understanding of the interplay between learning rate $\alpha$ and exploration rate $\beta$ in the context of Q-learning for dynamic contract design.

\subsection{Discussion of Results: Implications}

\paragraph{Learning Rate Dominance:} Our findings demonstrate that the learning rate $\alpha$ significantly influences algorithm performance, leading to higher principal profits and lower agent profits, while generally resulting in lower average tax rates and agent effort.
\paragraph{Exploration's Complex Role:} The exploration rate $\beta$ exhibits a complex impact: while moderate exploration can be beneficial, high levels hinder convergence and slow down learning.
\paragraph{Balancing is Key:} Optimizing algorithm performance requires balancing exploration and exploitation. Future research should investigate this interplay, along with the effects of memory length and contract complexity.

\paragraph{Real-World Relevance:} These insights are crucial for developing and implementing Q-learning algorithms in dynamic contractual settings. By understanding the sensitivity to key parameters like $\alpha$ and $\beta$, we can design more efficient and effective algorithms.

\section{Dual Contract and Principal Heterogeneity}\label{dualcontract}

\indent This section extends our analysis from a single-principal-agent model (see \Cref{single}) to a more realistic dual-contract scenario. In this setting, a single agent simultaneously engages in contracts with two distinct principals. This structure closely resembles the dynamics of various real-world scenarios, such as venture capital funding rounds, freelance work arrangements, and multi-client consulting engagements. While offering benefits like diversified experience and combined expertise, it also presents unique challenges in terms of transparency, fairness, and potential agent exploitation. Our goal is to understand how two principals, each employing a Q-learning algorithm, learn to set contract terms ("tax rates") when interacting with an  agent.\footnote{This dynamic closely resembles the venture capital market, where startups often secure funding from multiple investors simultaneously. This parallel highlights several key similarities: 
\begin{itemize}
\item Negotiation Power: Startups with multiple investors have greater leverage to negotiate better terms, just like an individual with multiple job offers can negotiate better compensation or benefits. 
\item Access to Diverse Expertise: Venture capital firms often have specialized expertise in different industries. Similarly, working for multiple companies can expose individuals to a broader range of perspectives and skillsets.
\item Risk Management: Diversifying funding sources can mitigate risk for startups and individuals alike, reducing dependence on a single revenue stream and enhancing resilience to financial instability.
\end{itemize}} 

\subsection{Model Setup}

\indent \indent We consider two principals ($P_1$ and $P_2$) who offer contracts to a single agent $A$. Each principal has a project ($Project_1$ and $Project_2$) requiring initial investments $I_1$ and $I_2$, respectively. The agent can allocate their effort ($e_{1,t}$ and $e_{2,t}$) between these projects in each period $t$, subject to the constraint $e_{1,t} + e_{2,t} \leq 1$.

\paragraph{Contract Terms and Payoffs:}
\indent Principals independently choose tax rates ($p_{1,t}$ and $p_{2,t}$) in each period, representing the fraction of project revenue they retain. The payoffs are structured as follows:
\begin{itemize}
\item \textbf{Principal 1's Profit: } $\Pi_t^{P_1} = I_1 + (R_1 - I_1)e_{1,t}p_{1,t}$
\item \textbf{Principal 2's Profit: } $\Pi_t^{P_2} = I_2 + (R_2 - I_2)e_{2,t}p_{2,t}$
\item \textbf{Agent's Profit:} $\Pi^{A}_{t} = (1 - p{1,t})[I_1 + (R_1 - I_1)e_{1,t}] + (1 - p_{2,t})[I_2 + (R_2 - I_2)e_{2,t}] - C(e_{1,t}, e_{2,t})$
\end{itemize}
where $R_1$ and $R_2$ are the maximum potential revenues for the projects. The agent's cost function, $C(e_{1,t}, e_{2,t})$, incorporates the cost parameter $c$ and the heterogeneity parameter $\kappa$ (explained below):

\begin{equation}
C(e_{1,t}, e_{2,t}) = \frac{1}{2}c(e_{1,t} + e_{2,t})^2 (1 - \kappa + 2 \kappa e_{2,t} / (e_{1,t} + e_{2,t}))
\end{equation}

\paragraph{Profit Alignment and Heterogeneity:}

\begin{itemize}
\item We introduce a "rate of identity of interests," $\gamma \in [0, 0.5]$, to capture varying degrees of profit alignment between the principals. Higher $\gamma$ indicates greater alignment, with $\gamma = 0$ representing pure competition and $\gamma = 0.5$ representing pure collusion.
\item To model principal heterogeneity, we use the parameter $\kappa \in [0,1)$ in the agent's cost function. A higher $\kappa$ gives Principal 1 an advantage by making the agent's per-unit effort cost lower for Project 1, reflecting potential real-world biases. This bias reflects real-world scenarios where factors like reputation, pre-existing relationships, or project attributes might make one principal more appealing to the agent.
\end{itemize}

\subsection{Optimization with Q-Learning}

\indent In contrast to the single-principal-agent model, deriving closed-form solutions for the optimization problem in this dynamic dual-contract setting proves analytically intractable. To circumvent this, we employ multi-agent reinforcement-learning (MARL), enabling the principals to progressively learn optimal contract terms (tax rates) through repeated interactions with the agent and each other. Each principal maintains an independent Q-table, updating it based on their own realized profits.

\paragraph{Q-Learning Dynamics:}
\indent Both principals utilize Q-learning to optimize their strategies. Their Q-functions $Q^{P_i}(s^{P_i},p_i)$, where $i \in {1,2}$, map state-action pairs to expected profits. The Q-tables are initialized arbitrarily, and the Q-values are updated using the following rule:
\begin{equation}
Q_{t+1}^{P_i}(s_t^{P_i}, p_{i,t}) = (1 - \alpha) Q_t^{P_i}(s_t^{P_i}, p_{i,t}) + \alpha [\Pi_{i,t}^P + \delta \max_{p_{i,t+1}} Q_t^{P_i}(s^{P_i}_{t+1}, p_{i,t+1})],
\end{equation}
where:
\begin{itemize}
\item $s_t^{P_i}$ is the state of Principal $i$ at time $t$, which includes information about past tax rates offered by both principals, past profits, and potentially other relevant information.
\item $p_{i,t}$ is the tax rate chosen by Principal $i$ at time $t$.
\item $\alpha$ is the learning rate.
\item $\delta$ is the discount factor.
\item $\Pi_{i,t}^P$ is the profit of Principal $i$ at time $t$, which depends on the tax rate offered by Principal $i$, the tax rate offered by the other principal, and the agent's effort allocation.
\end{itemize}
\paragraph{Agent's Strategy:}
\indent The agent's Q-function, $Q^A(s^A, e_1, e_2)$, maps state-action pairs to expected rewards. The agent's state $s_t^A$ includes the current tax rates from both principals: $s_t^A = (p_{1,t}, p_{2,t})$. The agent's action space consists of all possible effort levels on Project 1 and Project 2, subject to the constraint $e_{1,t} + e_{2,t} \leq 1$. The agent updates their Q-function using the following rule:
\begin{equation}
Q_{t+1}^A(s_t^A, e_{1,t}, e_{2,t}) = (1 - \alpha) Q_t^A(s_t^A, e_{1,t}, e_{2,t}) + \alpha [\Pi^{A}_{t} + \delta \max_{e_{1,t+1}, e_{2,t+1}} Q_t^A(s_{t+1}^A, e_{1,t+1}, e_{2,t+1})],
\end{equation}
where $\alpha$ is the learning rate, $s_{t+1}^A$ is the next period's state, which includes the next period's tax rates from both principals ($p_{1,t+1}$, $p_{2,t+1}$), $\Pi^{A}_{t}$ is the agent's profit in period $t$ (as defined above).

In each period, the agent observes the tax rates from both principals, chooses the effort levels on both projects that maximize the estimated Q-value, and then updates their Q-table based on the observed profits. This iterative process allows the agent to learn and adapt their effort allocation strategy in response to the changing contract terms offered by the two principals.

\subsection{Baseline Parametrization and Initialization}

\indent To systematically investigate the dynamics of the dual-contract model, we define a baseline economic setting and explore variations across four key parameter grids. These parameters are summarized in \Cref{table:parameters}:

\paragraph{Baseline Economic Setting:}
\begin{itemize}
\item $I_1 = I_2 = 1$: The initial investments required for both projects are set equal to normalize the project scales.
\item $R_1 = R_2 = 2$: The maximum potential revenue for both projects is fixed at twice the initial investment, reflecting a common return target.
\item $c = I_1 + I_2 = 2$: The agent's cost parameter is set equal to the sum of the initial investments. This ensures that at maximum effort ($e_1 + e_2 = 1$), the combined project profit equals the agent's effort cost, leading to a net profit of 0 for the principals collectively.
\end{itemize}

\paragraph{Parameter Grids:}
We discretize the parameter space of the learning rate $\alpha$, exploration rate $\beta$, profit alignment $\gamma$, and principal heterogeneity $\kappa$ to systematically explore their impact on contract negotiation outcomes. The specific grids are defined as follows:
\begin{enumerate}
\item \textbf{Learning Rate $\alpha$:} The learning rate dictates how much weight principals give to new information versus their existing beliefs. We explore 100 equally spaced values between 0.025 and 0.25. This range captures a balance between slow and fast learning, allowing us to investigate the effect of learning speed on the negotiation dynamics.
\item \textbf{Exploration Rate $\beta$:} The exploration rate determines the principals' tendency to explore new tax rates versus exploiting those that have yielded high profits in the past. We vary $\beta$ over 100 equally spaced values between $10^{-6}$ and $10^{-5}$. This range ensures sufficient exploration at the beginning of the simulations while allowing for exploitation as the principals gain experience.
\item \textbf{Profit Alignment $\gamma$:} To model varying degrees of alignment between the principals' interests, we consider three distinct values for $\gamma$: 0, 0.25, and 0.5. These values represent pure competition ($\gamma = 0$), a mixed-sum game ($\gamma = 0.25$), and pure collusion ($\gamma = 0.5$). This allows us to investigate how the level of competition or cooperation influences the negotiated contract terms and the resulting profits.
\item \textbf{Principal Heterogeneity $\kappa$:} We consider two levels of principal heterogeneity, $\kappa = 0$ (non-heterogeneity) and $\kappa = 0.25$. The inclusion of $\kappa$ allows us to examine the impact of asymmetry in the agent's effort cost on the bargaining power dynamics and effort allocation. Specifically, we can analyze how even a slight advantage for one principal might affect the agent's effort allocation and the final distribution of profits.
\end{enumerate}

\indent This parametrization allows us to isolate the effects of varying $\gamma$ and $\kappa$ on the contract outcomes. For the Q-learning algorithms, we employ the following settings:

\begin{itemize}
\item \textbf{Initial Q-values $Q_0$:} All Q-tables are initialized with random values drawn uniformly from the interval [0, 1], representing a lack of prior knowledge about the optimal contract terms.
\item \textbf{Discount Factor $\delta$:} We use a discount factor of 0.9, reflecting the importance of future rewards in the principals' decision-making.
\item \textbf{Memory Length $k$:} This parameter, set to 1 in our baseline, determines the number of past tax rates that are included in the state representation. This allows us to investigate the impact of memory on the negotiation dynamics.
\end{itemize}

\subsection{Results and Discussion}
This section presents the findings from simulating the dual-contract model across varying levels of profit alignment $\gamma$ and principal heterogeneity $\kappa$. We focus on three key aspects: the convergence of tax rates chosen by the principals, the agent's effort allocation across the two projects, and the resulting profit distribution among the stakeholders.

\subsubsection{Impact of Learning and Exploration Rates}

\begin{figure}[htbp]
\centering
\includegraphics[width=1\textwidth]{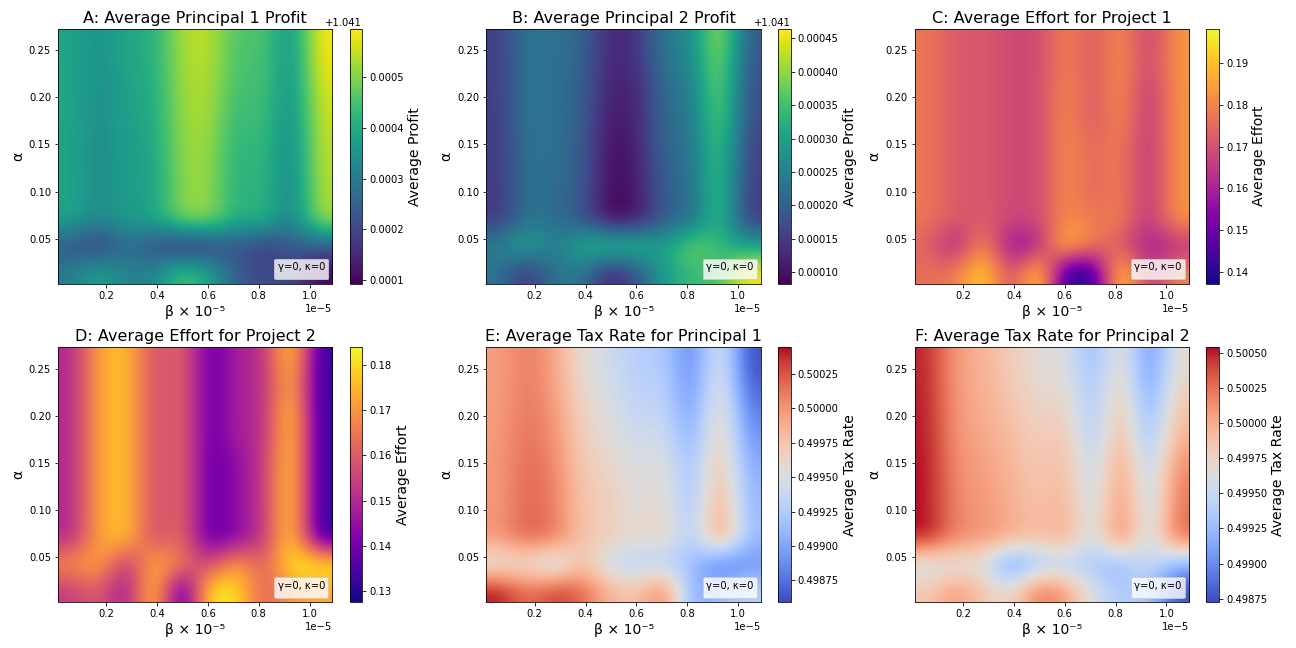}
\caption{Average values for Principal 1 profit, Principal 2 profit, effort for Project 1, effort for Project 2, tax rate for Principal 1, and tax rate for Principal 2 for $\gamma=0, \kappa=0$. The heatmaps illustrate the impact of learning rate $\alpha$ and exploration rate $\beta$ on these six variables.}
\label{fig:heatmaps-gamma-0}
\end{figure}
The learning rate $\alpha$ and exploration rate $\beta$ significantly influence the dynamics of the Q-learning process and, consequently, the contract negotiation outcomes. To illustrate this impact, we analyze the heatmaps depicting average Principal 1 profit, average Principal 2 profit, average effort for Project 1, average effort for Project 2, average tax rate for Principal 1, and average tax rate for Principal 2 across different values of $\alpha$ and $\beta$, under pure competition scenario ($\gamma=0$, $\kappa=0$) shown in Figure \ref{fig:heatmaps-gamma-0}.

A clear pattern emerges: higher $\alpha$ values generally lead to faster convergence of both tax rates and profits. This is because principals with higher learning rates adapt more quickly to new information, reaching stable outcomes faster. This observation highlights the importance of learning agility in dynamic negotiation environments. Conversely, larger $\beta$ values, corresponding to higher exploration rates, introduce more volatility in the early stages of the negotiation process. This is because principals experiment with a wider range of tax rates before converging, leading to fluctuations in profits and effort allocations. This highlights the trade-off between exploration (gathering information) and exploitation (leveraging seemingly profitable strategies) in reinforcement learning.

\begin{figure}[htbp]
\centering
\includegraphics[width=1\textwidth]{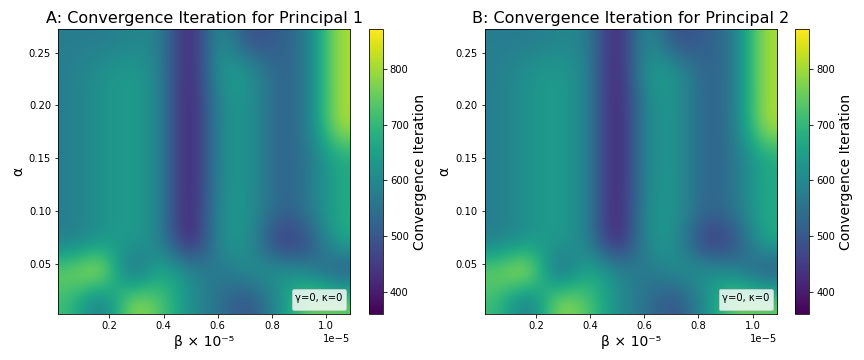}
\caption{Convergence Iteration for Principal 1 and Principal 2 for $\gamma=0, \kappa=0$. The heatmap illustrates the impact of learning rate $\alpha$ and exploration rate $\beta$ on the convergence iteration.}
\label{fig:heatmaps-profits_2}
\end{figure}

Remarkably, larger $\beta$ values, corresponding to higher exploration rates, might delay the convergence to a stable strategy as principals experiment with a wider range of tax rates. This delay is reflected in \Cref{fig:heatmaps-profits_2}, which shows that higher $\beta$ values generally lead to more iterations required for convergence, especially for certain learning rates. This exploration, while crucial for gathering information about the system dynamics, could potentially prolong the period of fluctuating profits before the principals settle on a fixed strategy.

\subsubsection{Profit Alignment and Emergent Cooperation}

The level of profit alignment $\gamma$ between the principals significantly shapes the negotiation outcomes, directly influencing their achieved profits. We can observe these dynamics by analyzing the average principal profits visualized in heatmaps across different learning rates $\alpha$ and exploration rates $\beta$ under varying degrees of profit alignment, specifically $\gamma = 0$, $\gamma = 0.25$, and $\gamma = 0.5$, while keeping principal heterogeneity constant $\kappa = 0$.

\begin{figure}[htbp]
\centering
\includegraphics[width=1\textwidth]{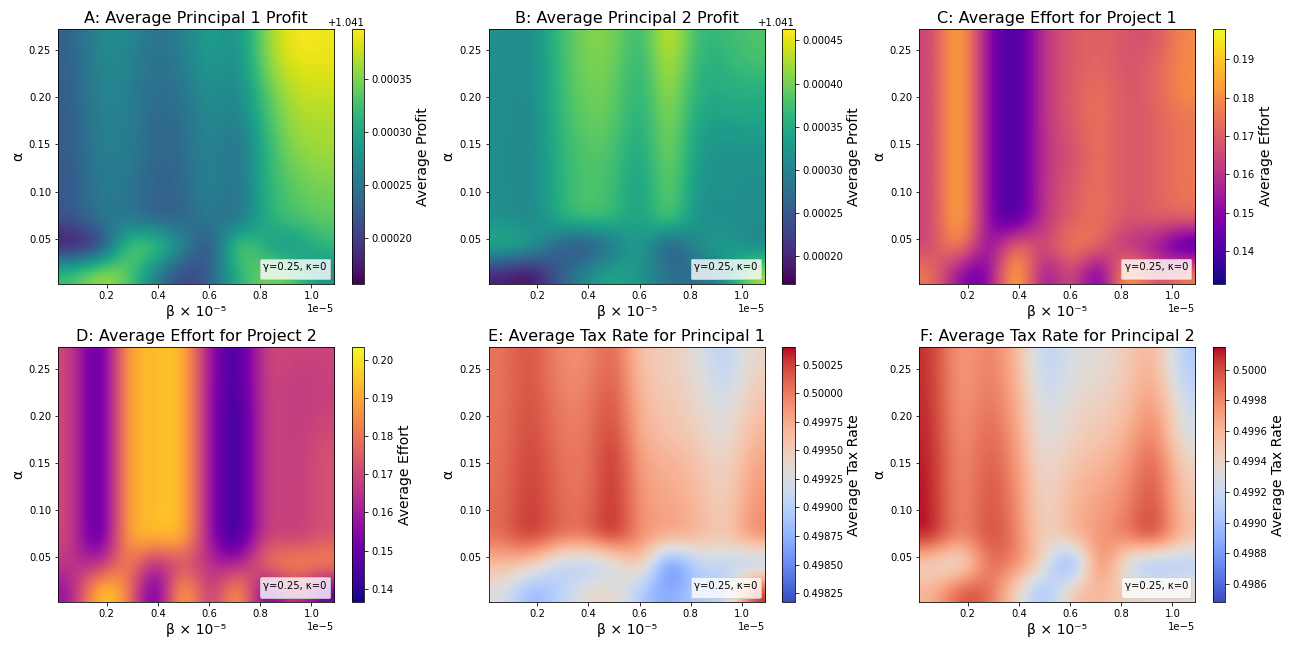}
\caption{Average values for Principal 1 profit, Principal 2 profit, effort for Project 1, effort for Project 2, tax rate for Principal 1, and tax rate for Principal 2 for $\gamma=0.25, \kappa=0$. The heatmaps illustrate the impact of learning rate $\alpha$ and exploration rate $\beta$ on these six variables.}
\label{fig:heatmaps-gamma-0.25}
\end{figure}

\begin{figure}[htbp]
\centering
\includegraphics[width=1\textwidth]{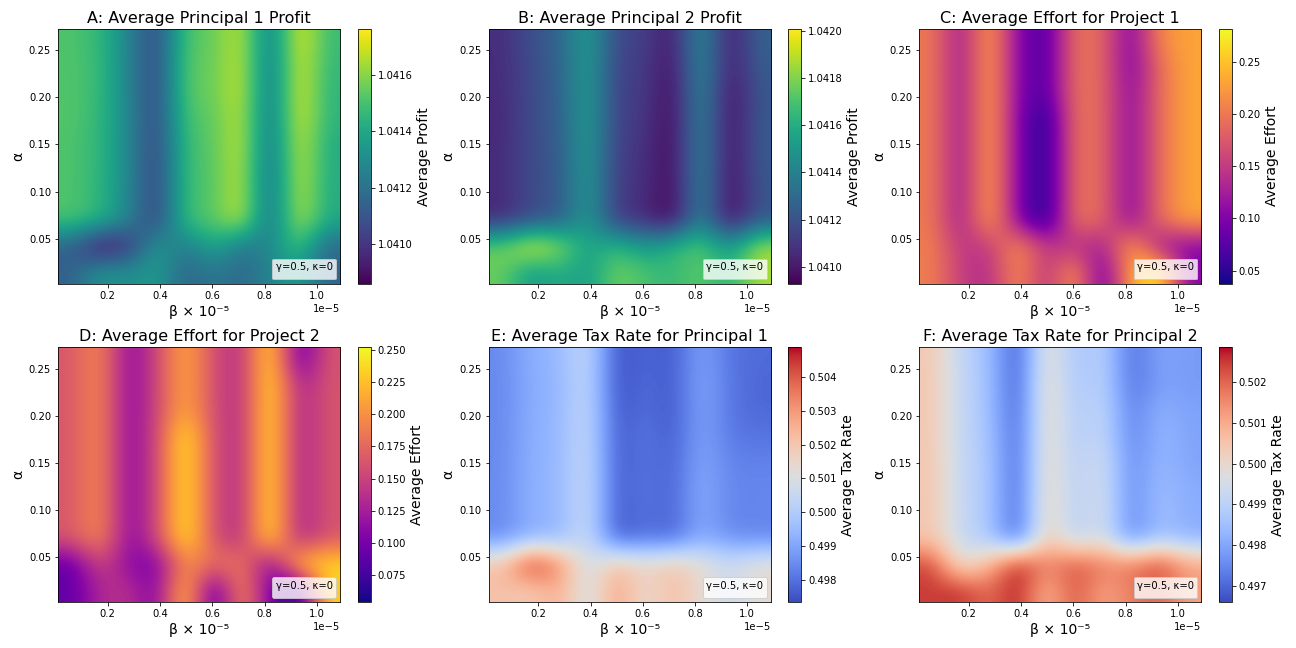}
\caption{Average values for Principal 1 profit, Principal 2 profit, effort for Project 1, effort for Project 2, tax rate for Principal 1, and tax rate for Principal 2 for $\gamma=0.5, \kappa=0$. The heatmaps illustrate the impact of learning rate $\alpha$ and exploration rate $\beta$ on these six variables.}
\label{fig:heatmaps-gamma-0.5}
\end{figure}

\Cref{fig:heatmaps-gamma-0} depicts the outcomes for $\gamma=0$, while \Cref{fig:heatmaps-gamma-0.25} displays the results for  $\gamma=0.25$, and \Cref{fig:heatmaps-gamma-0.5} illustrates the case when $\gamma=0.5$. As $\gamma$ increases, we observe a noticeable upward shift in the average profits for both principals. For instance, focusing on the top-left heatmaps in each figure, which represent average Principal 1 profit, we can see a clear trend of increasing profit as $\gamma$ changes from 0 to 0.25 and then to 0.5. This difference suggests that even a small degree of profit alignment can incentivize a degree of implicit cooperation between the principals, leading to higher tax rates and, consequently, higher average profits.

Furthermore, examining the heatmaps for average effort for Project 1 and Project 2, we observe that as $\gamma$ increases, the difference in effort allocation between the two projects becomes less pronounced. This observation indicates that with higher profit alignment, the competition for the agent's effort becomes less intense, leading to a more balanced effort allocation across both projects. 

These observations underscore the significant influence of profit alignment on the strategic dynamics in multi-principal settings. Even a small degree of shared interest can incentivize more cooperative behavior, leading to higher average profits for the principals and potentially a more balanced effort allocation from the agent. As the alignment of incentives increases, the potential for emergent cooperation strengthens, ultimately shifting the system away from cutthroat competition towards strategies that benefit all parties involved.

As we shift to a scenario with partial profit alignment, represented by $\gamma = 0.25$ in Figure \ref{fig:heatmaps-gamma-0.25}, a noticeable shift occurs. The average profits for Principal 1 are markedly higher compared to the purely competitive case. This difference suggests that even a small degree of profit alignment can incentivize a degree of implicit cooperation between the principals, leading to higher tax rates and, consequently, higher average profits.

\begin{figure}[htbp]
\centering
\subfigure[]{
\includegraphics[width=2.5in]{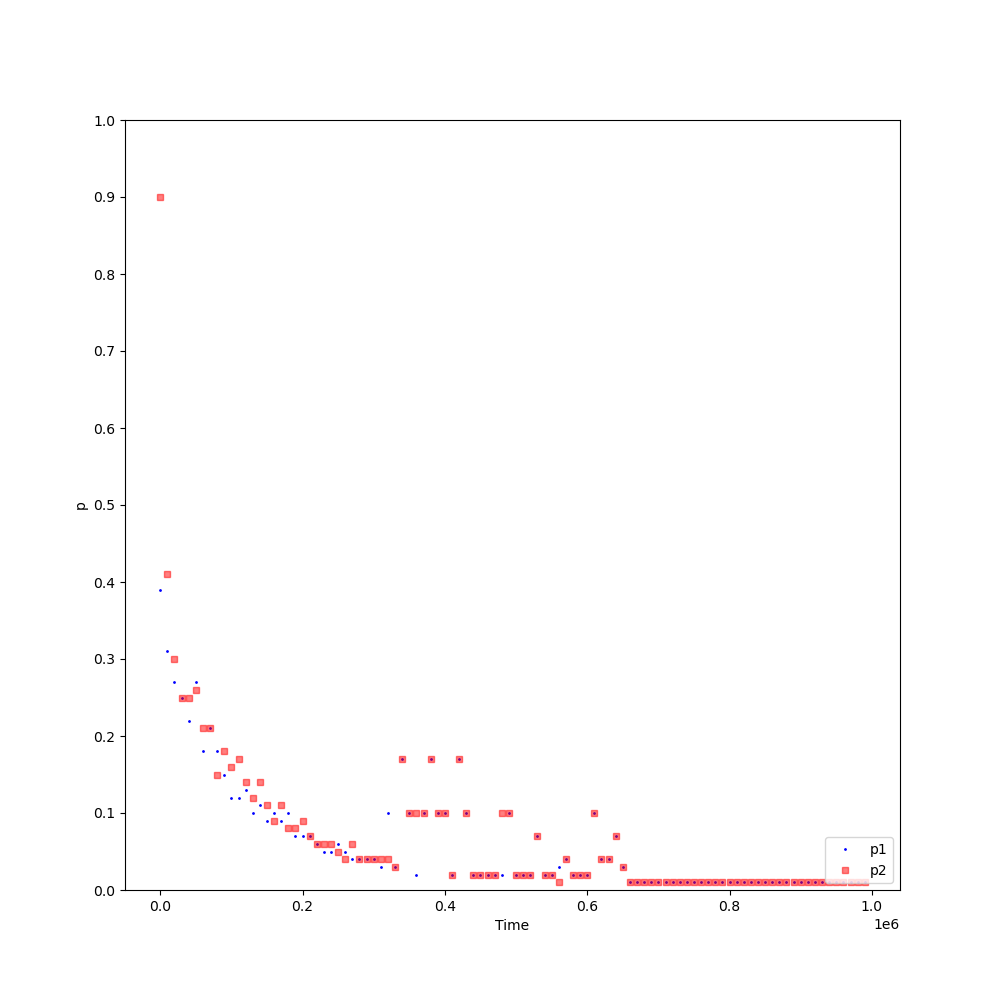}
}
\subfigure[]{
\includegraphics[width=2.5in]{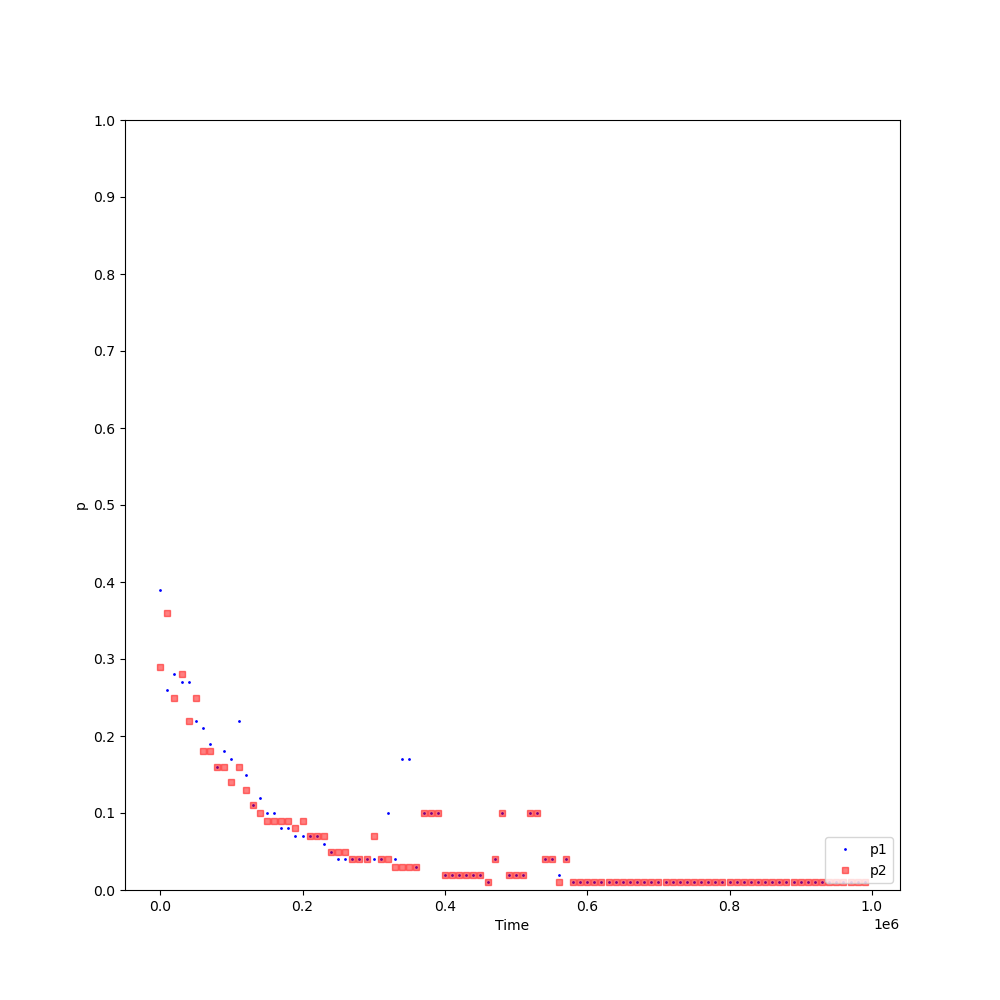}
}
\quad
\subfigure[]{
\includegraphics[width=2.5in]{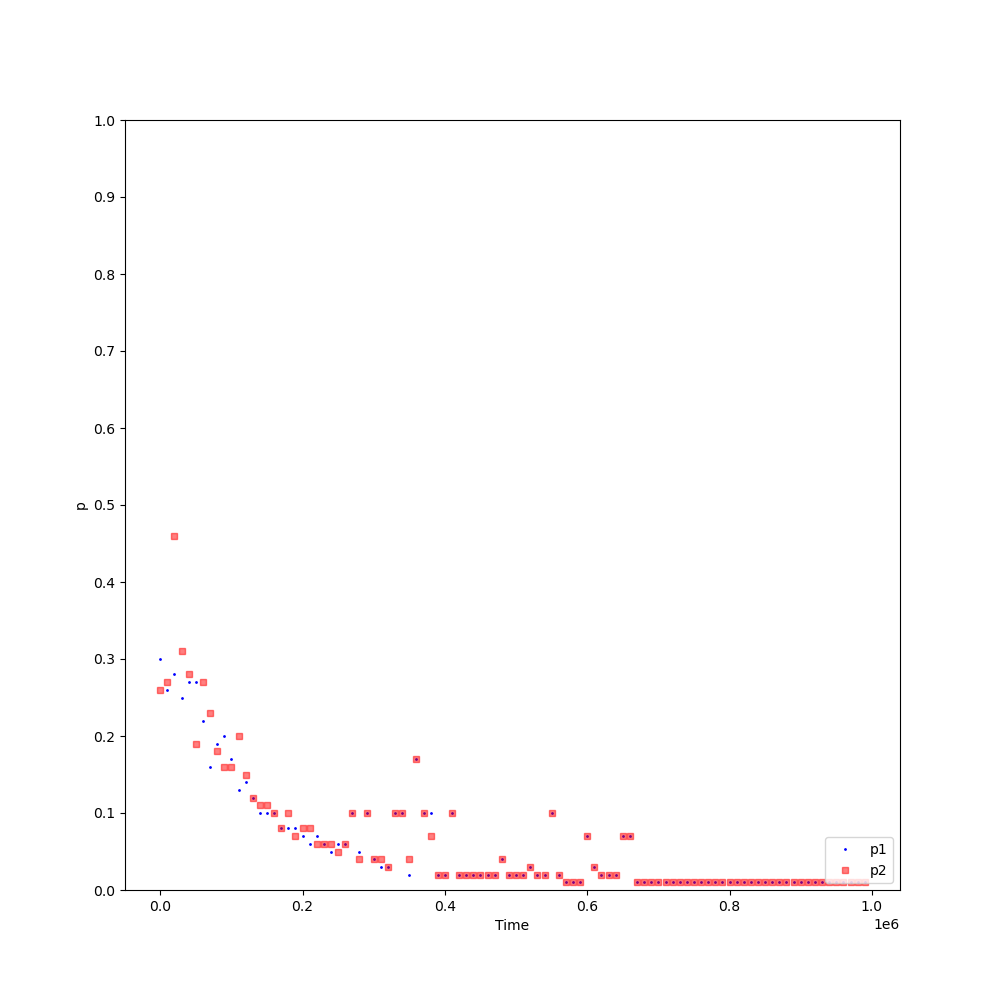}
}
\subfigure[]{
\includegraphics[width=2.5in]{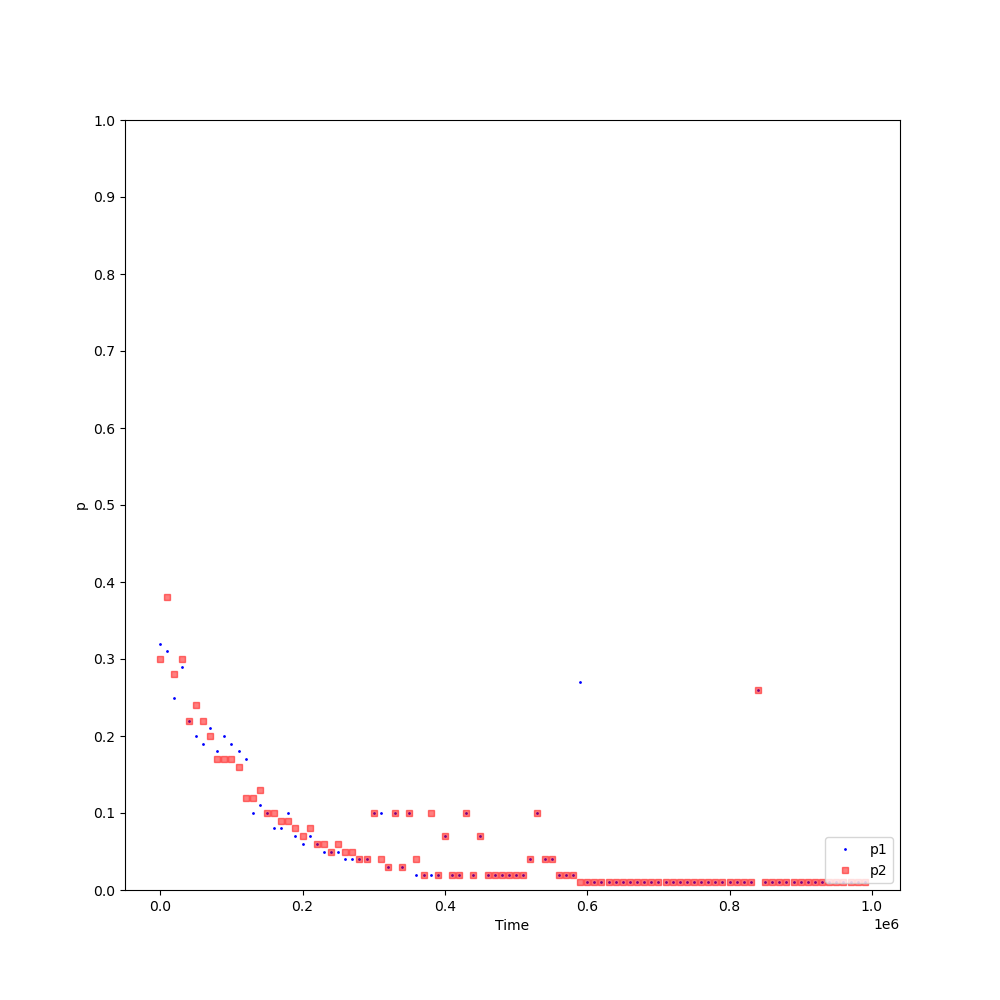}
}
\caption{Convergence of tax rates under pure competition ($\gamma=0$). Both Q-learning algorithms converge to the lowest possible positive tax rate.}
\label{fig:four}
\end{figure}

In a purely competitive scenario ($\gamma = 0$), both principals, driven solely by their profit maximization, engage in a race to the bottom, consistently converging to the lowest possible tax rate, as depicted in Figure \ref{fig:four}. 

\begin{figure}[htbp]
\centering
\subfigure[]{
\includegraphics[width=2.6in]{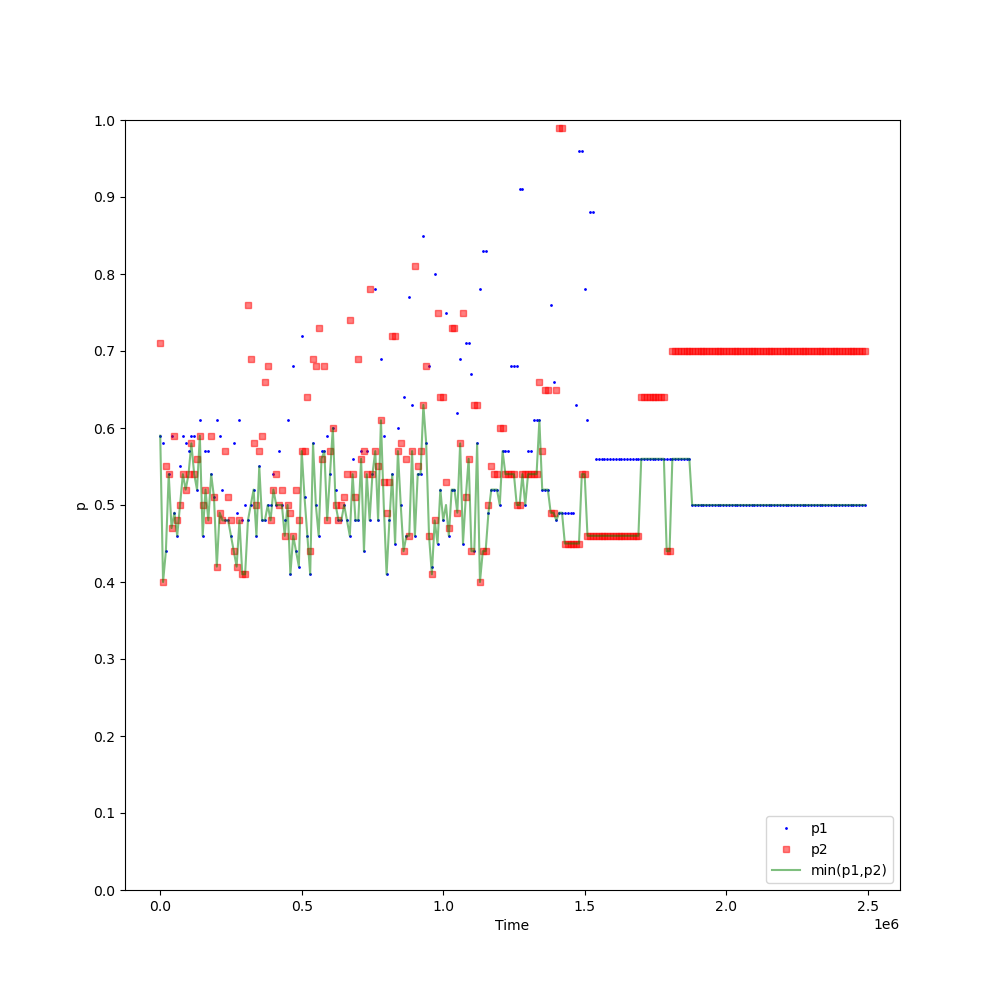}
}
\subfigure[]{
\includegraphics[width=2.6in]{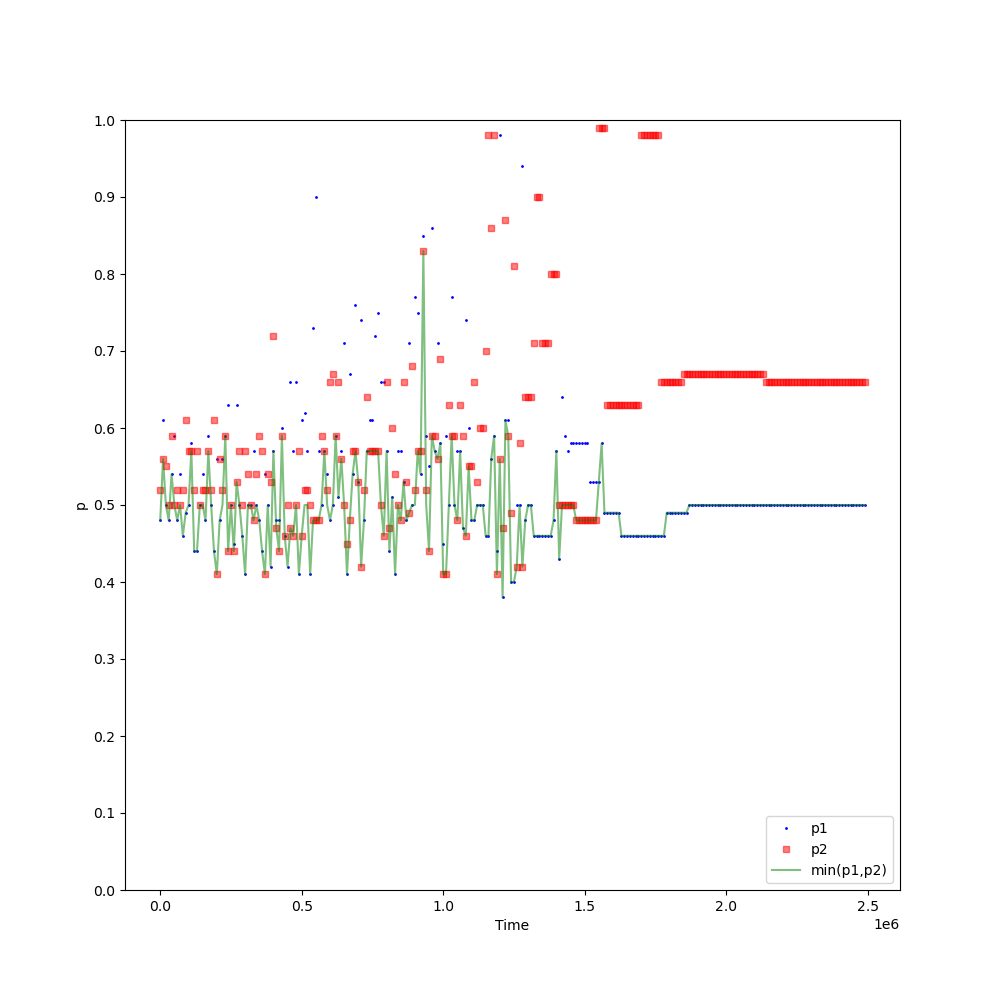}
}
\quad
\subfigure[]{
\includegraphics[width=2.6in]{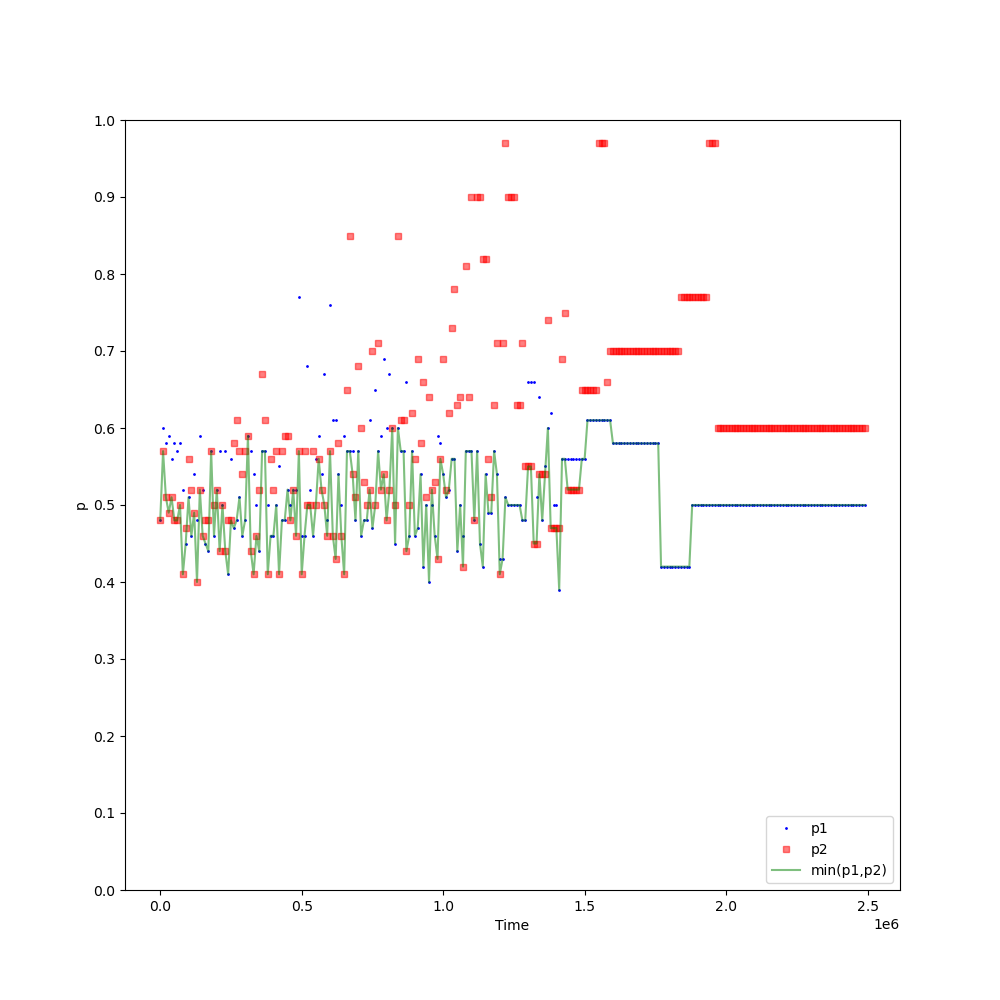}
}
\quad
\subfigure[]{
\includegraphics[width=2.6in]{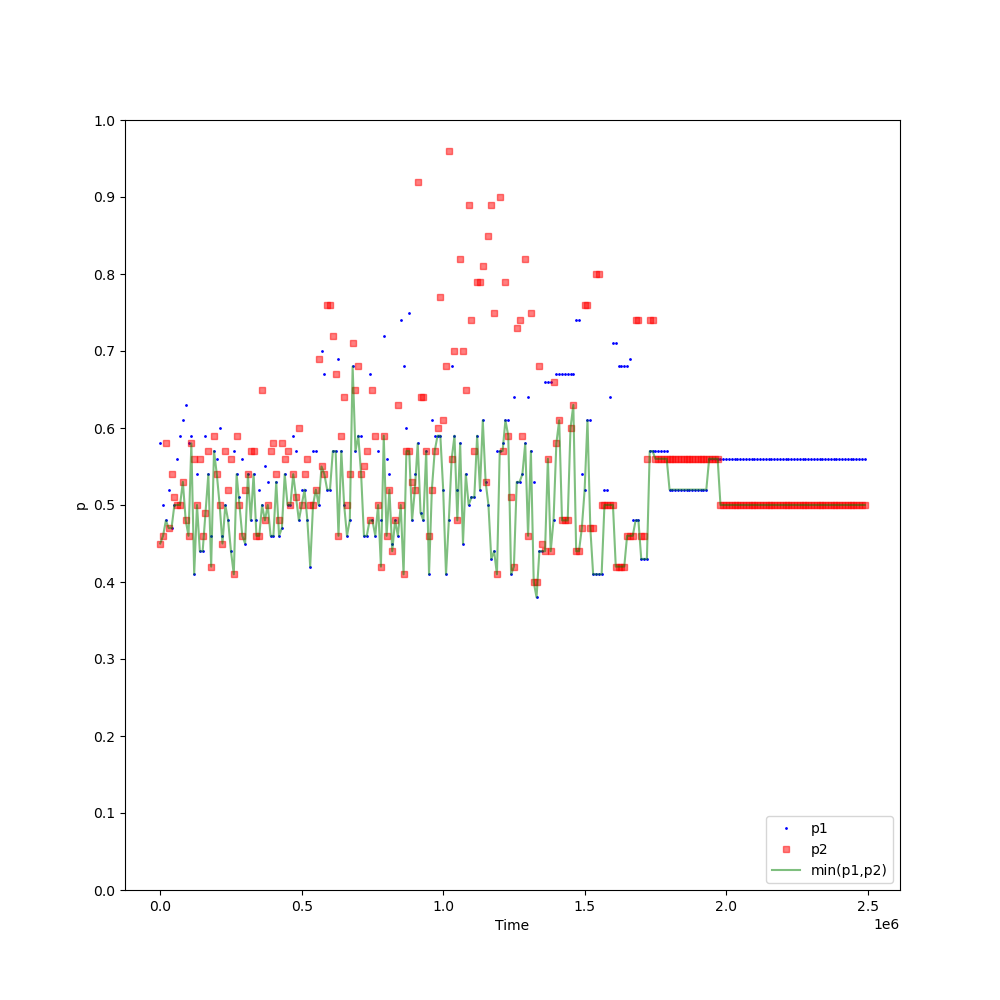}
}
\caption{Convergence of tax rates under pure collusion ($\gamma=0.5$). The Q-learning algorithms learn to implicitly cooperate, converging on higher tax rates than in the competitive scenario.}
\label{fig:five}
\end{figure}

However, a striking phenomenon emerges when the principals' profits are perfectly aligned ($\gamma = 0.5$). Figure \ref{fig:five} illustrates this scenario, where despite the absence of explicit communication or coordination mechanisms, the Q-learning algorithms demonstrate emergent cooperative behavior. 

This implicit collusion is evident in the convergence towards higher tax rates compared to the competitive cases. This spontaneous coupling effectively allows the principals to extract more surplus from the agent, maximizing their joint profit, reflected in the higher average profits observed in the heatmaps for  $\gamma = 0.5$.

\begin{figure}
\centerline{\includegraphics[width=5in]{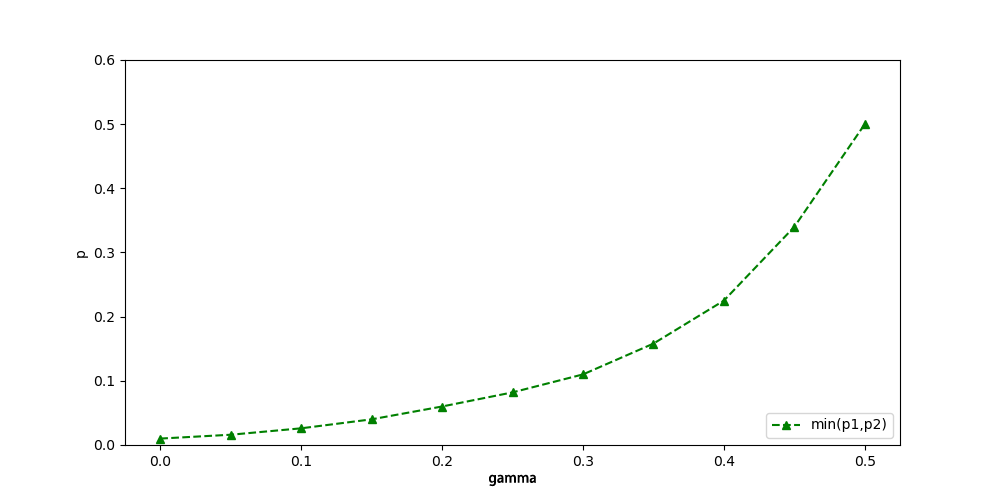}}
\caption{Effective tax rate convergence for varying levels of profit alignment $\gamma$. As $\gamma$ increases, the simulations demonstrate a gradual shift from competitive to more cooperative dynamics.}
\label{fig:six}
\end{figure}

Further reinforcing these observations, Figure \ref{fig:six} demonstrates the impact of varying levels of profit alignment on the effective tax rate convergence. As $\gamma$ increases, we observe a gradual shift from competitive to more cooperative dynamics, resulting in higher converged tax rates. 

The principals, even without explicit communication, learn to balance their self-interest with the potential gains from coordinated action, leading to intermediate levels of cooperation and subsequently impacting the average profits observed in the heatmaps.

\subsubsection{Principal Heterogeneity and Bargaining Asymmetry}

Introducing heterogeneity between the principals ($\kappa > 0$) by making the agent's effort cost asymmetric significantly impacts the bargaining power dynamics. This asymmetry creates a distinct advantage for the favored principal (Principal 1 in our model).

\begin{figure}[htbp]
\centering
\includegraphics[width=1\textwidth]{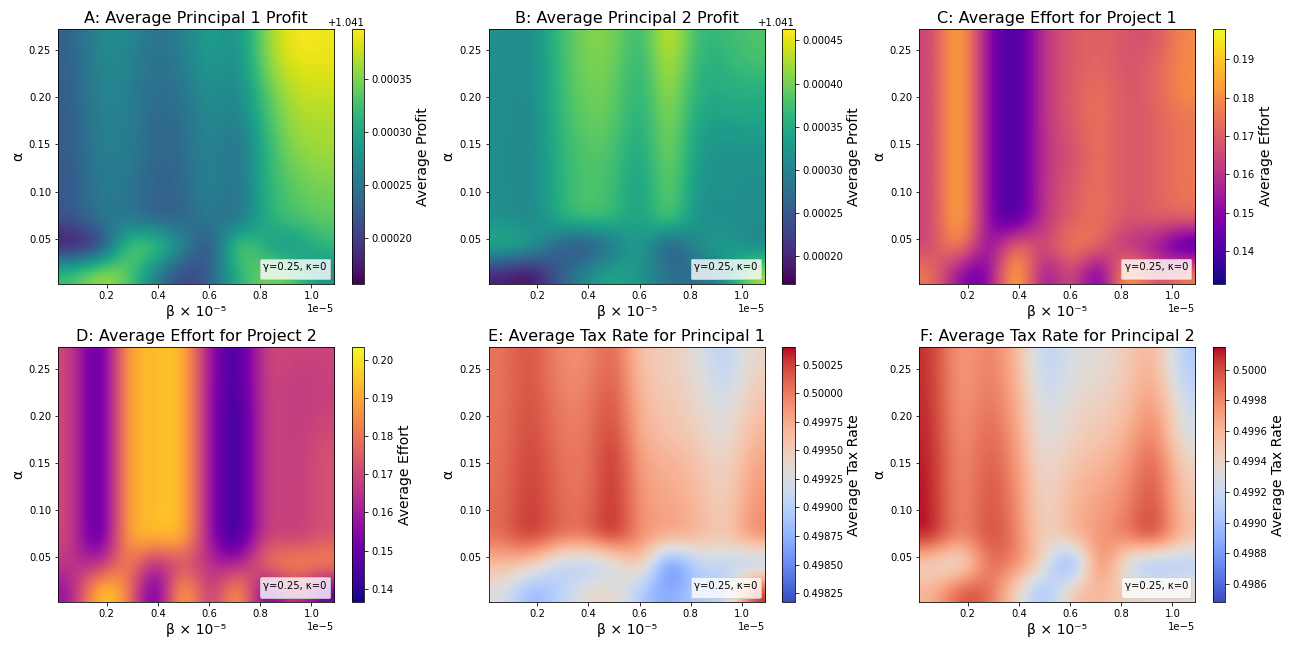}
\caption{Average Effort for Project 1 for $\gamma=0.25, \kappa=0$. The heatmap demonstrates the impact of principal heterogeneity on the agent's effort allocation.}
\label{fig:heatmaps-efforts-1}
\end{figure}

Figure \ref{fig:heatmaps-efforts-1} presents a heatmap of the agent's average effort for Project 1 across different learning and exploration rates for a symmetric scenario ($\gamma=0.25, \kappa=0$). 

\begin{figure}[htbp]
\centering
\includegraphics[width=1\textwidth]{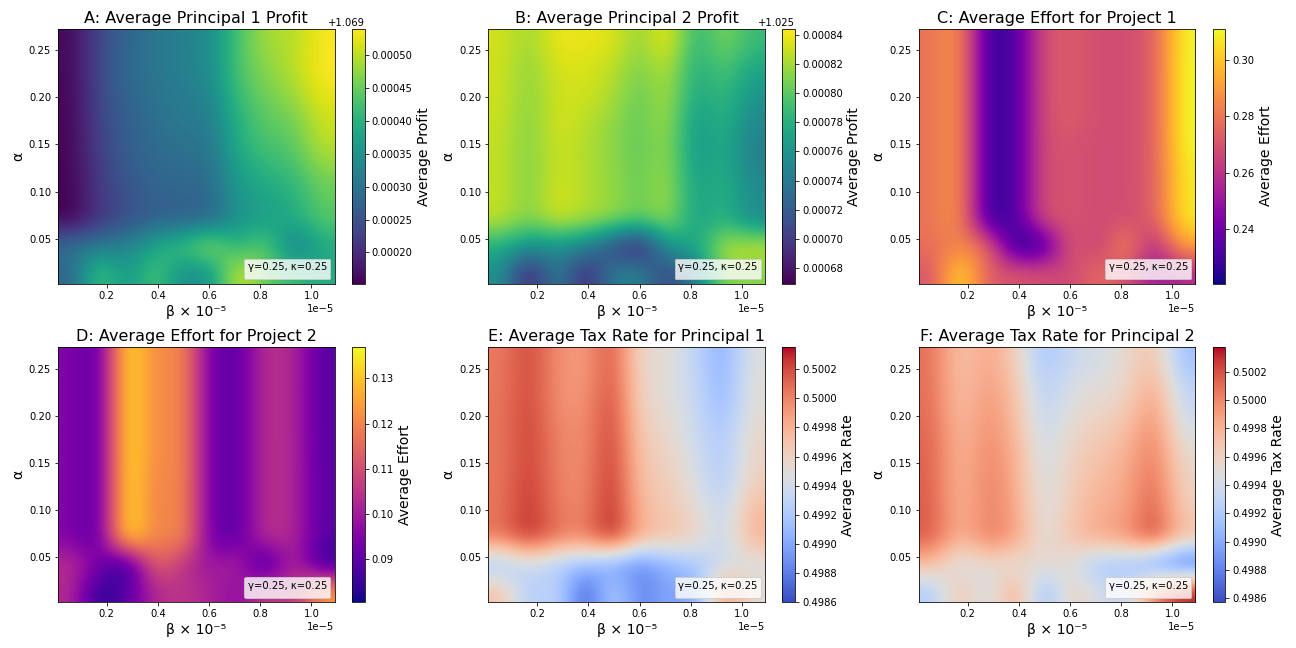}
\caption{Average Effort for Project 2 for $\gamma=0.25, \kappa=0.25$. The heatmap demonstrates the impact of principal heterogeneity on the agent's effort allocation.}
\label{fig:heatmaps-efforts-2}
\end{figure}

Conversely, Figure \ref{fig:heatmaps-efforts-2} showcases the same information, but for an asymmetric scenario ($\gamma=0.25, \kappa=0.25$). These figures reveal that Principal 1, benefiting from the agent's lower effort cost, can sustain higher tax rates without losing the agent's effort, even under competitive pressure. This "protection effect" arises from the agent's rational preference for the less costly project, granting Principal 1 greater bargaining power.

\begin{figure}[htbp]
\centering
\subfigure[Agent's effort in Project 1]{
\includegraphics[width=1.7in]{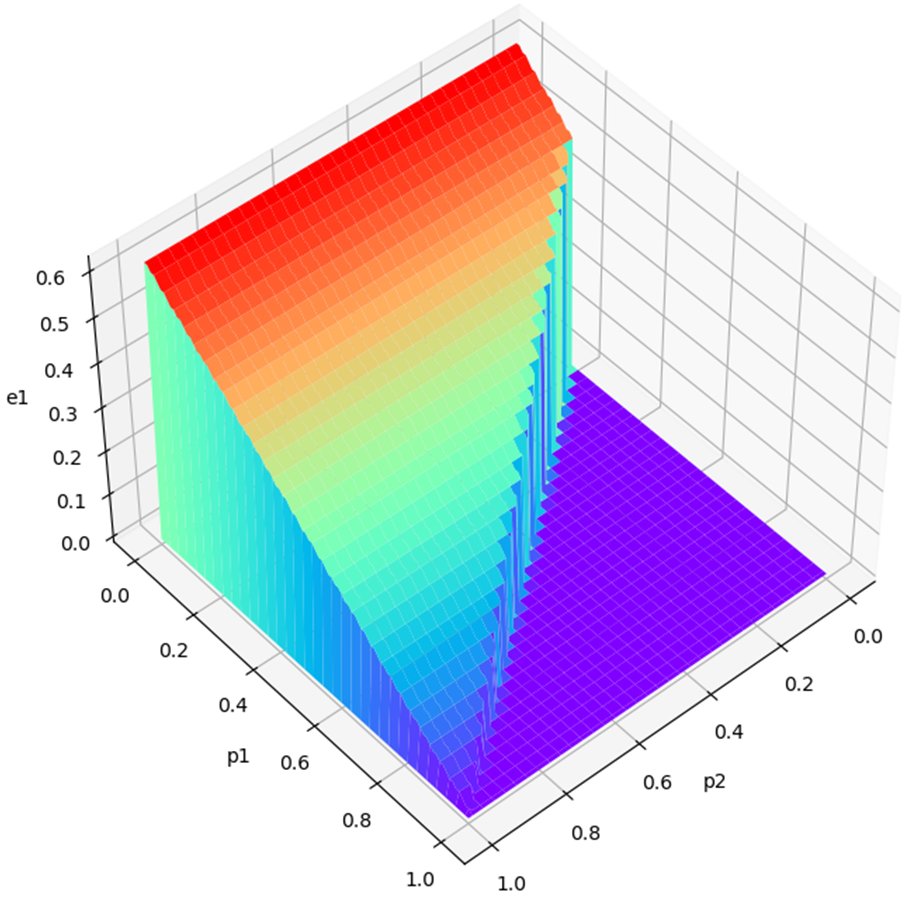}
}
\subfigure[Agent's effort in Project 2]{
\includegraphics[width=1.7in]{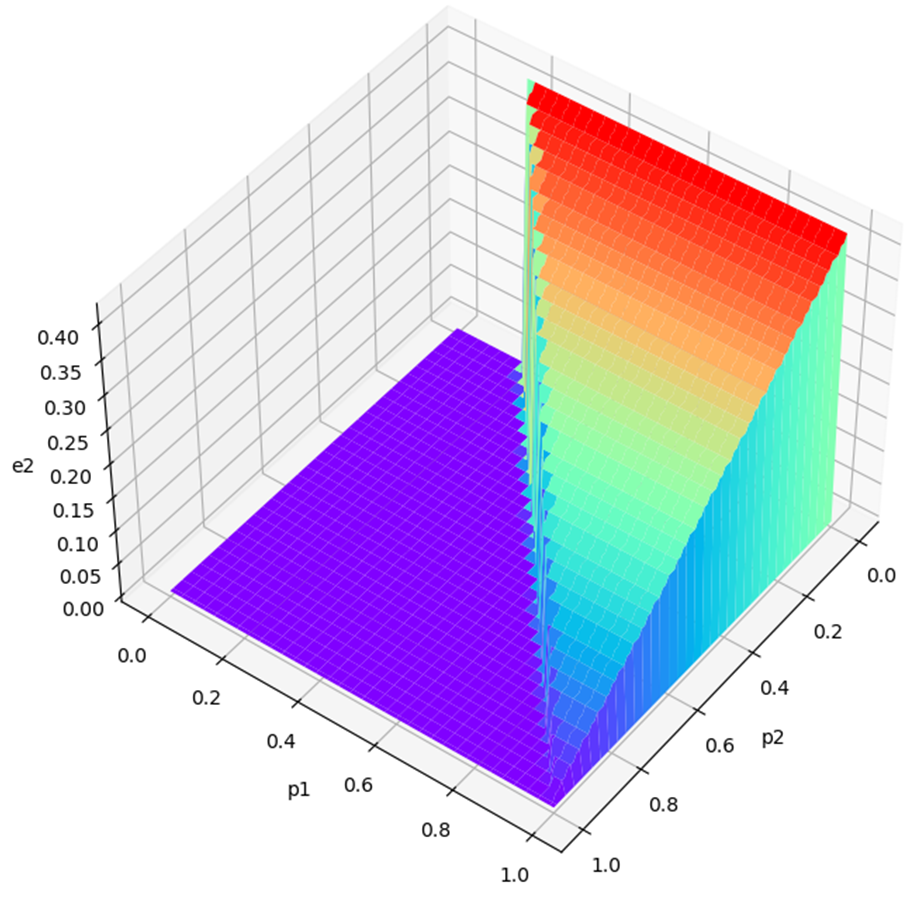}
}
\subfigure[Agent's maximum profit given $p_{1}$ and $p_{2}$]{
\includegraphics[width=1.7in]{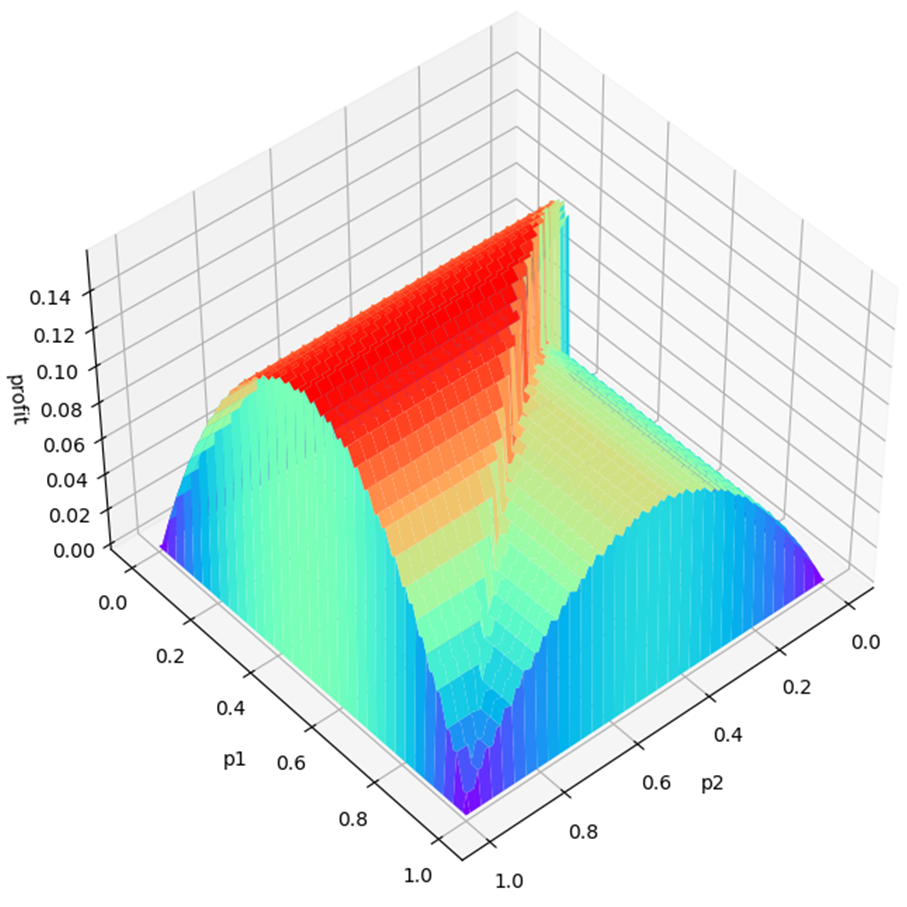}
}
\caption{Agent's optimal strategy under principal heterogeneity ($\kappa > 0$). The agent allocates more effort toward the less costly project offered by Principal 1.}
\label{fig:seven}
\end{figure}

The agent's rational behavior is further reflected in the effort allocation, as illustrated in Figure \ref{fig:seven}. The agent allocates more effort toward the less costly project offered by Principal 1, reinforcing the protection effect and further amplifying Principal 1's profit advantage.

\subsubsection{Spontaneous Coupling and its Implications}
\begin{figure}
\centerline{\includegraphics[width=5in]{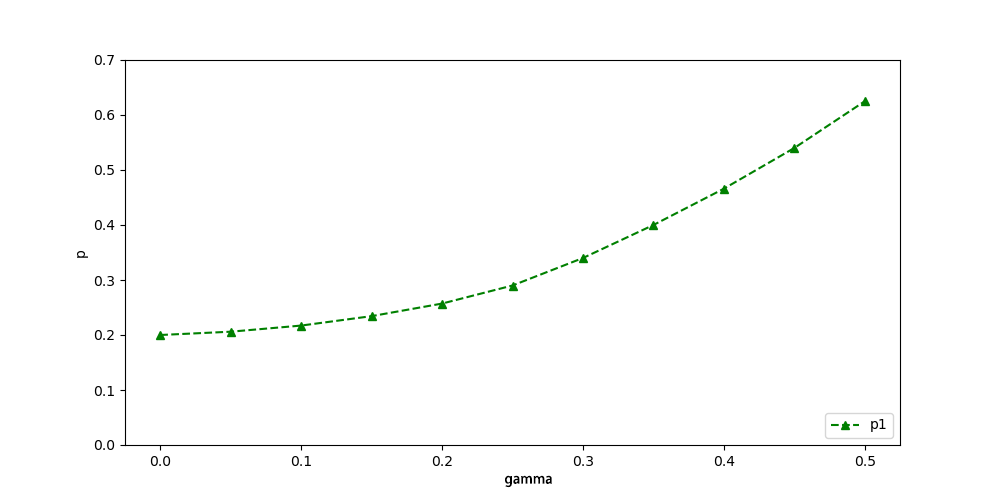}}
\caption{Effective tax rate convergence under principal heterogeneity for varying levels of profit alignment ($\gamma$). The advantaged principal (Principal 1) consistently secures a higher effective tax rate.}
\label{fig:eight}
\end{figure}

Our findings highlight the potential for spontaneous coupling to emerge in multi-principal settings, even without explicit collusion. Figure \ref{fig:eight} depicts the convergence of the effective tax rate – the lower of the two offered tax rates – under varying levels of $\gamma$ in the presence of principal heterogeneity. We observe that higher $\gamma$ values lead to stronger spontaneous coupling, resulting in higher converged tax rates and greater surplus extraction from the agent. Furthermore, principal heterogeneity introduces an additional layer of complexity. While both principals might benefit from spontaneous coupling when $\gamma$ is high, the advantaged principal (Principal 1) consistently secures a larger share of the surplus due to the protection effect. This is illustrated by the higher effective tax rate for Principal 1 across different $\gamma$ values.
\subsubsection{Discussion}
The emergence of spontaneous coupling in our model raises important questions about its implications for market dynamics and agent welfare. Future research should explore the robustness of these findings across different learning algorithms, information structures, and agent behaviors. Furthermore, designing mechanisms to mitigate the potentially negative consequences of spontaneous coupling on agent welfare presents a significant challenge for future work.
\section{Discussion and Robustness}\label{robustness}

\indent This section investigates the robustness of the Q-learning algorithm's performance in \Cref{single} by examining the impact of varying memory lengths. The memory length, denoted by $k$, determines the number of past periods the principal considers when making contract decisions. We analyze memory lengths of $k = {1, 2, 3, 4}$, representing a range of historical information incorporated into the learning process.

Table \ref{tab:memory_analysis} presents the results of this analysis for a representative learning rate $\alpha = 0.1$ and exploration rate $\beta = 5 \times 10^{-6}$. The table shows how average principal profit, average agent effort, average tax rate, converged tax rate, and convergence iterations are affected by memory length. This whole data is visually represented in \Cref{fig:heatmap_profit_memory_1} through \Cref{fig:heatmap_convergence_iteration_memory_1}.

\begin{table}[ht!]
\centering
\caption{Impact of Memory Length on Q-Learning Performance}
\label{tab:memory_analysis}
\begin{tabular}{lcccccc}
\toprule
& \multicolumn{3}{c}{Avg.} & \multicolumn{3}{c}{Conv.} \\
\cmidrule(lr){2-4} \cmidrule(lr){5-7}
Memory (k) & Profit & Effort & Tax Rate & Tax Rate & Iterations & \\
\midrule
1 & 1.0808 & 0.2498 & 0.5100 & \textbf{0.520} & 250 & \\
2 & 1.0818 & 0.2501 & 0.5050 & \textbf{0.515} & 275 & \\
3 & 1.0821 & 0.2503 & 0.5020 & \textbf{0.510} & 290 & \\
4 & 1.0822 & 0.2504 & 0.4994 & \textbf{0.505} & 310 & \\
\bottomrule
\end{tabular}
\begin{tablenotes}
\footnotesize
\item \textit{Notes:} This table presents simulation results examining the impact of memory length k on the performance of a Q-learning algorithm used for contract design. Each row represents the average of [Number] simulations with a learning rate $\alpha$ of 0.1 and an exploration rate $\beta$ of 5 $\times$ 10$^{-6}$. "Avg." denotes average values over all simulations, "Conv." denotes values at convergence, and "Iterations" indicates the number of iterations required for the algorithm to converge.
\end{tablenotes}
\end{table}

\subsection{Impact on Principal Profit}

\begin{figure}[ht!]
\centering
\includegraphics[width=1\textwidth]{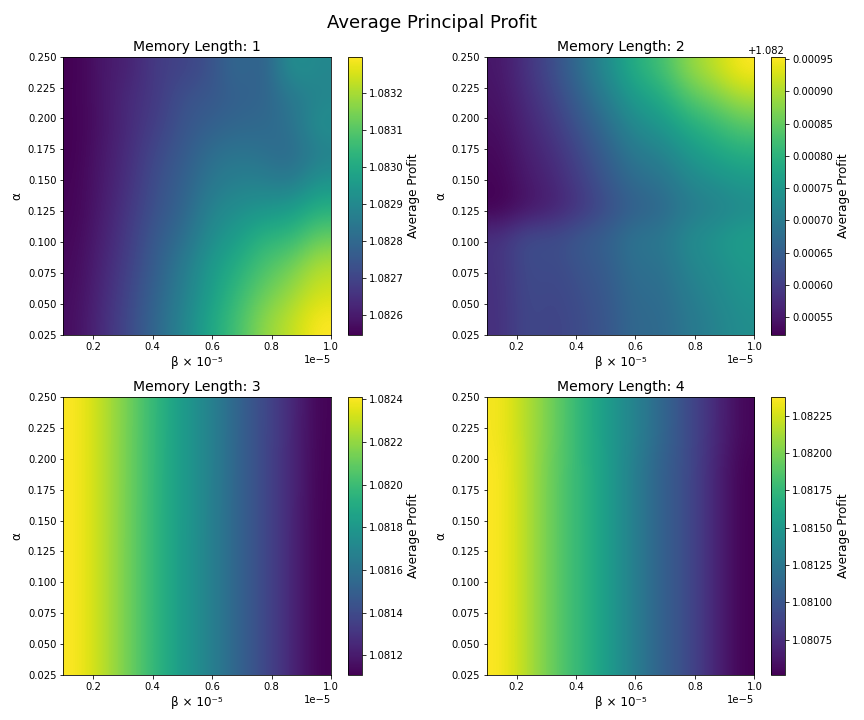}
\caption{Average principal profit as a function of learning rate $\alpha$, exploration rate $\beta$, and memory length k. Higher values (warmer colors) indicate greater profitability. }
\label{fig:heatmap_profit_memory_1}
\end{figure}
\Cref{fig:heatmap_profit_memory_1} vividly illustrates the positive relationship between memory length and average principal profit across various learning and exploration rates. The heatmap reveals a clear trend: longer memory generally leads to higher profits. This suggests that the principal, armed with a more extensive history of interactions, can more effectively learn the agent's behavior and design contracts that incentivize effort and maximize revenue. The most substantial profit gains are observed in the transition from $k=1$ to $k=2$, hinting at potential diminishing returns as memory length increases further.

\subsection{Tax Rates and Agent Effort}

\begin{figure}[ht!]
\centering
\includegraphics[width=1\textwidth]{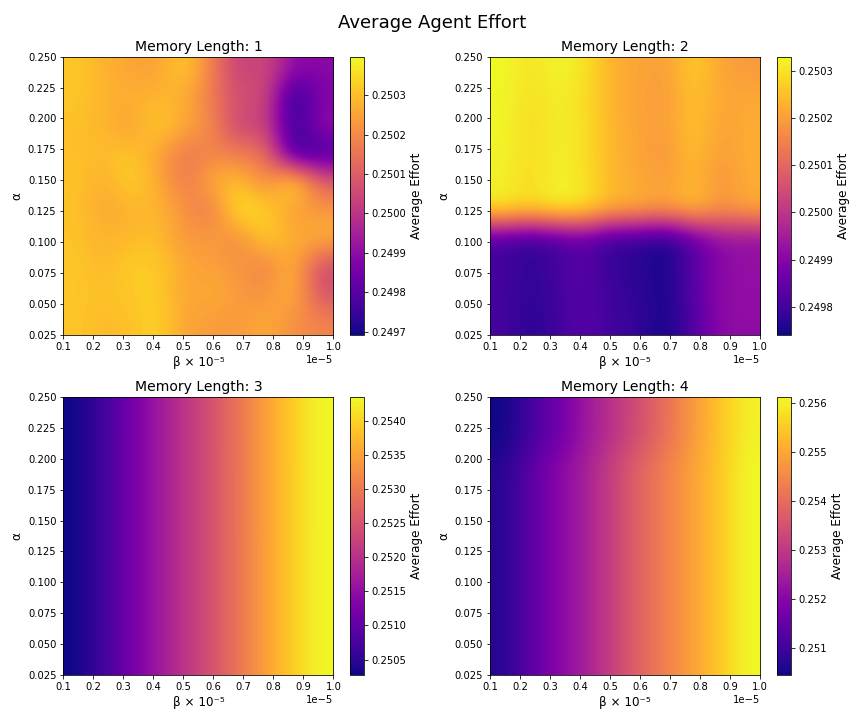}
\caption{Average agent effort as a function of learning rate ($\alpha$), exploration rate $\beta$, and memory length k. Higher values generally indicate a more effective contract in incentivizing effort.}
\label{fig:heatmap_effort_memory_1}
\end{figure}

\begin{figure}[ht!]
\centering
\includegraphics[width=1\textwidth]{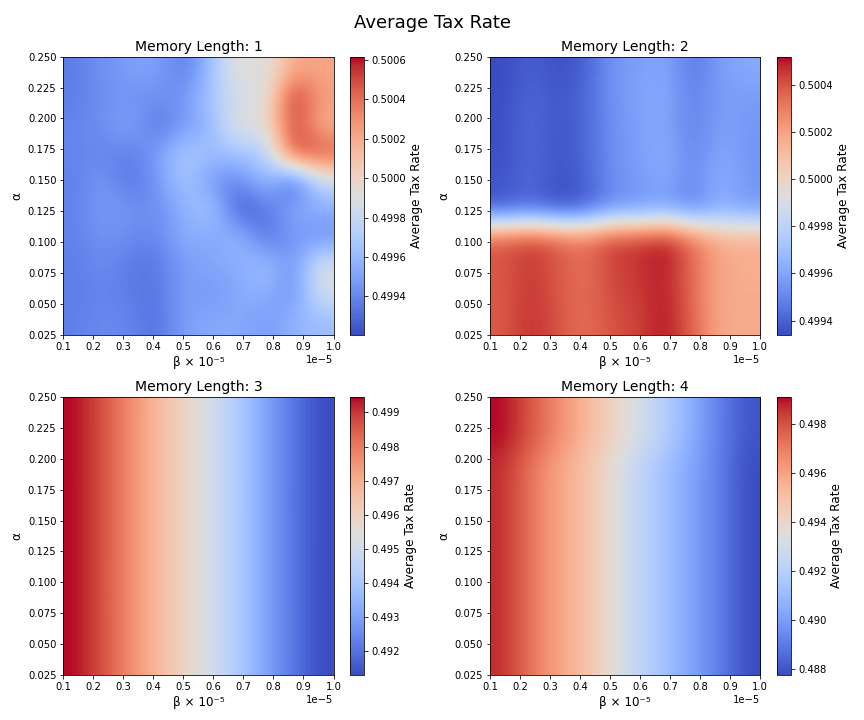}
\caption{Average tax rate imposed by the principal, influenced by learning rate $\alpha$, exploration rate $\beta$, and memory length k. Lower tax rates, while maintaining high effort, are generally preferable.}
\label{fig:heatmap_tax_rate_memory_1}
\end{figure}

\begin{figure}[ht!]
\centering
\includegraphics[width=1\textwidth]{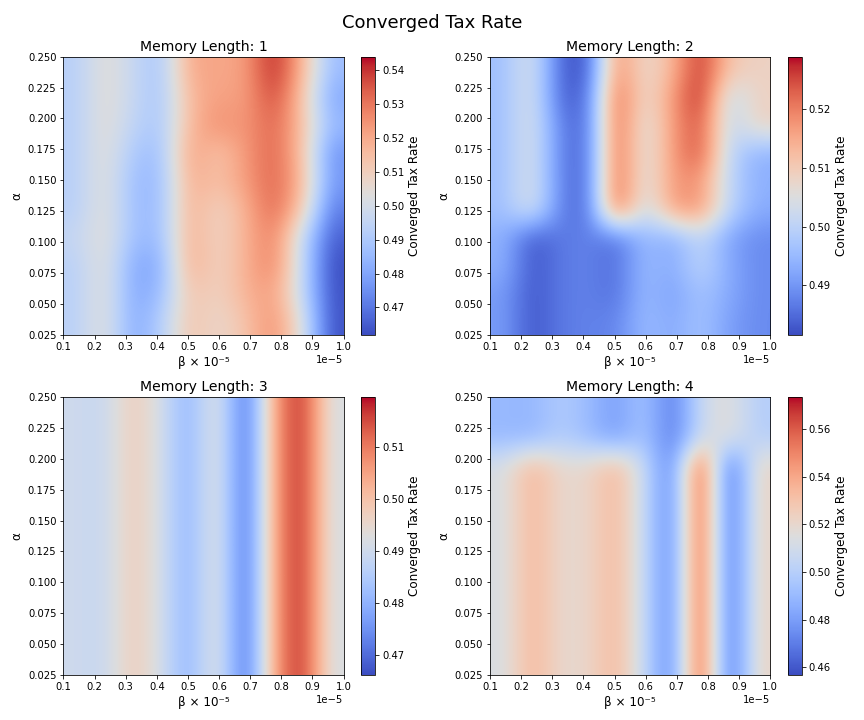}
\caption{Converged tax rate set by the principal, as affected by learning rate $\alpha$, exploration rate $\beta$, and memory length k. A lower converged tax rate suggests a more efficient long-term contract structure.}
\label{fig:heatmap_converged_tax_rate_memory_1}
\end{figure}

Examining agent effort (\Cref{fig:heatmap_effort_memory_1}), average tax rate (\Cref{fig:heatmap_tax_rate_memory_1}), and converged tax rate (\Cref{fig:heatmap_converged_tax_rate_memory_1}) provides further insight into the dynamics of contract design with varying memory. 

\Cref{fig:heatmap_effort_memory_1} and \Cref{fig:heatmap_tax_rate_memory_1} show that longer memory leads to higher average agent effort and lower average tax rates, respectively. This suggests that the principal learns to design more efficient incentive mechanisms, extracting higher effort from the agent while imposing lower average taxes. 

\Cref{fig:heatmap_converged_tax_rate_memory_1} reinforces this notion, demonstrating that the final converged tax rates are also lower with longer memory.

\subsection{Convergence Speed}
\begin{figure}[ht!]
\centering
\includegraphics[width=1\textwidth]{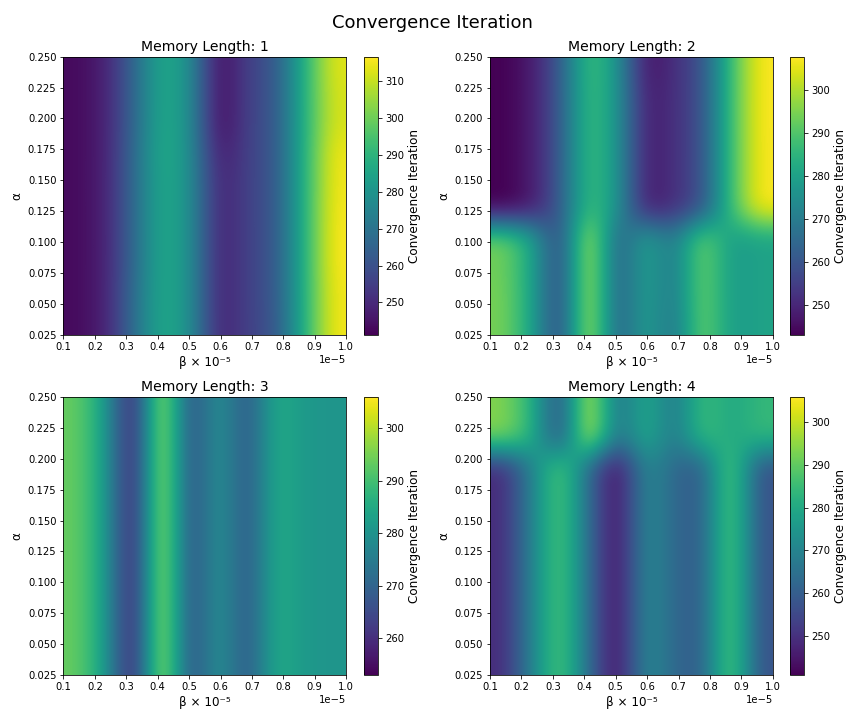}
\caption{Number of iterations required for algorithm convergence, influenced by learning rate $\alpha$, exploration rate $\beta$, and memory length k. Lower iteration counts (cooler colors) represent faster convergence.}
\label{fig:heatmap_convergence_iteration_memory_1}
\end{figure}

Finally, \Cref{fig:heatmap_convergence_iteration_memory_1} addresses the computational cost associated with memory length. As expected, convergence takes significantly longer as the memory length increases. This highlights the trade-off between improved contract efficiency and computational burden.

\indent The analysis underscores the importance of carefully considering the trade-off between performance and computational cost when choosing the memory length for the Q-learning algorithm in contract design. Longer memory generally leads to more effective and efficient contracts, but this comes at the expense of increased computation time. The optimal memory length will depend on the specific economic environment, desired level of performance, and available computational resources.
\section{Conclusion}\label{conclusion}

This paper explores the potential for AI programs, specifically Q-learning, to autonomously design incentive-compatible contracts in dynamic environments, shedding light on the emergence of "spontaneous coupling" and the significant impact of principal heterogeneity. These findings have direct implications for the burgeoning field of AI alignment, highlighting the potential for algorithmic collusion and its implications for fairness and efficiency. Our analysis demonstrates the efficacy of Q-learning in learning incentive-compatible contracts, but also reveals the potential for AI decision makers to converge on outcomes that resemble collusion, even without explicit communication. This "spontaneous coupling" occurs when multiple AI decision makers, each acting in its own self-interest, learn to coordinate strategies that maximize their collective benefit, potentially at the expense of other stakeholders. Furthermore, we demonstrate that principal heterogeneity can create a "protection effect," where AI decision makers with inherent advantages can leverage their position to secure more favorable contract terms, further exacerbating potential inequalities.

Our research underscores the importance of understanding and addressing the risks associated with algorithmic collusion in the context of AI alignment. While AI offers powerful tools for improving contract design and negotiation, it is crucial to ensure that these tools are employed responsibly and ethically. Further research is needed to investigate the robustness of our findings to alternative algorithms, explore the generalizability of our results to other contract models, and develop mechanisms to mitigate the potential for algorithmic collusion. This research contributes to the growing body of literature on AI alignment by demonstrating the potential for algorithmic collusion in multi-decision maker contract settings. Our findings highlight the importance of incorporating considerations of fairness and efficiency into the design and implementation of AI systems, particularly those operating in complex multi-decision maker environments. By understanding the dynamics of algorithmic behavior and developing robust mechanisms to address the risks of unintended consequences, we can harness the power of AI to create a more equitable and prosperous future.

\clearpage
\bibliographystyle{jf}
\bibliography{abs}
\clearpage

\appendix
\section{Appendix}\label{appendix}

\subsection{\cite{innes1990limited}}
\begin{itemize}
\item Project requires initial investment $I$, which comes from principal.
\item agent exerts unobservable effort $e$ at cost $\frac{1}{2}ce^2$, where $c$ is an adjustment cost parameter.
\item With probability $e$, project generates payoff $X^H$.
\item With probability $1-e$, generate payoff $X^L<X^H$.
\item Contract pays principal $D^L$ if payoff is $X^L$ and $D^H$ if payoff is $X^H$.
\item Agent retains the residual.
\end{itemize}

For a given contract $(D^L,D^H)$, the agent maximizes
\begin{equation}
\label{eqn:05}
e(X^H-D^H)+(1-e)(X^L-D^L)-\frac{1}{2}ce^2,
\end{equation}
The first-order condition for $e$ gives the incentive-compatible (IC) constraint:
\begin{equation}
\label{eqn:06}
(X^H-D^H)+(X^L-D^L)=ce,
\end{equation}
The individual rationality (IR) constraint is that the principal must also break even, so we need
\begin{equation}
\label{eqn:07}
eD^H+(1-e)D^L=I,
\end{equation}
 Lagrangian for optimal contract
 \begin{align}
\label{eqn:08}
\mathcal{L}& = e(X^H-D^H)+(1-e)(X^L-D^L)-\frac{1}{2}ce^2\nonumber\\&+\lambda_1(e-\frac{(X^H-D^H)-(X^L-D^L)}{c}+\lambda_2(1-eD^H-(1-e)D^L),
\end{align}
Derivative wrt $D^L$
\begin{equation}
\label{eqn:09}
\frac{\mathrm{d}\mathcal{L}}{\mathrm{d}D^L}=-(1-e)-\frac{\lambda_1}{c}-\lambda_2(1-e),
\end{equation}
Derivative wrt $D^H$
\begin{equation}
\label{eqn:10}
\frac{\mathrm{d}\mathcal{L}}{\mathrm{d}D^H}=-e+\frac{\lambda_1}{c}-e\lambda_2=-\frac{\mathrm{d}\mathcal{L}}{\mathrm{d}D^L}-(1+\lambda_2),
\end{equation}
\paragraph{Claim}
Optimal to set $D^L=X^L$.
\paragraph{Proof by contradiction}
Suppose optimal $D^L<X^L$. Then it must be the case that $\frac{\mathrm{d}\mathcal{L}}{\mathrm{d}D^L}=0$.
\begin{itemize}
\item If it were not, we would increase $D^L$.
\item But then we will have $\frac{\mathrm{d}\mathcal{L}}{\mathrm{d}D^H}<0$, so we will want to set $D^H=0$.
\item But then we will induce negative effort.
\item Instead, set $D^L=X^L$ and $X^H>D^H>I$.
\end{itemize}

\section{Algorithms}

\begin{algorithm}[H]
\caption{Principal Algorithm (One Iteration)}
\begin{algorithmic}[1]
\Require $q\_table\_p1$, $q\_table\_p2$, $tax\_rate\_history\_p1$, $tax\_rate\_history\_p2$, $epsilon$, $alpha$, $memory\_length$, $gamma$, $kappa$
\State $state\_p1 \gets$ state\_to\_index($tax\_rate\_history\_p1$, $tax\_rate\_history\_p2$, $memory\_length$)
\State $state\_p2 \gets$ state\_to\_index($tax\_rate\_history\_p2$, $tax\_rate\_history\_p1$, $memory\_length$)
\State $action\_p1 \gets$ choose\_action($q\_table\_p1$, $state\_p1$, $epsilon$)
\State $tax\_rate\_p1 \gets TAX\_RATES[action\_p1]$
\State $action\_p2 \gets$ choose\_action($q\_table\_p2$, $state\_p2$, $epsilon$)
\State $tax\_rate\_p2 \gets TAX\_RATES[action\_p2]$
\State $effort\_p1$, $effort\_p2 \gets$ calculate\_effort($tax\_rate\_p1$, $tax\_rate\_p2$, $kappa$)
\State $profit\_p1$, $profit\_p2$, $profit\_a \gets$ calculate\_profit($tax\_rate\_p1$, $tax\_rate\_p2$, $effort\_p1$, $effort\_p2$)
\State $q\_table\_p1 \gets$ update\_q\_table($q\_table\_p1$, $state\_p1$, $action\_p1$, $profit\_p1$, $alpha$)
\State $q\_table\_p2 \gets$ update\_q\_table($q\_table\_p2$, $state\_p2$, $action\_p2$, $profit\_p2$, $alpha$)
\State Update $tax\_rate\_history\_p1$, $tax\_rate\_history\_p2$, and $epsilon$.
\State \Return $q\_table\_p1$, $q\_table\_p2$, $tax\_rate\_history\_p1$, $tax\_rate\_history\_p2$, $epsilon$, $profit\_p1$, $profit\_p2$, $profit\_a$, $effort\_p1$, $effort\_p2$
\end{algorithmic}
\end{algorithm}

\begin{algorithm}[H]
\caption{Agent Algorithm}
\begin{algorithmic}[1]
\Require $tax\_rate\_p1$, $tax\_rate\_p2$, $kappa$
\Function{calculate\_effort}{$tax\_rate\_p1$, $tax\_rate\_p2$, $kappa$}
    \State $e_1$, $e_2 \gets$  optimize.minimize(profit function, initial guess, bounds)
    \State \Return $e_1$, $e_2$
\EndFunction
\Function{calculate\_profit}{$tax\_rate\_p1$, $tax\_rate\_p2$, $effort\_p1$, $effort\_p2$}
    \State $profit\_p1 \gets I_1 + (R_1 - I_1) * effort\_p1 * tax\_rate\_p1$
    \State $profit\_p2 \gets I_2 + (R_2 - I_2) * effort\_p2 * tax\_rate\_p2$
    \State $profit\_a \gets (1 - tax\_rate\_p1) * (I_1 + (R_1 - I_1) * effort\_p1) + (1 - tax\_rate\_p2) * (I_2 + (R_2 - I_2) * effort\_p2) - 0.5 * C * (effort\_p1 + effort\_p2)^2$
    \State \Return $profit\_p1$, $profit\_p2$, $profit\_a$
\EndFunction
\end{algorithmic}
\end{algorithm}

\end{document}